\documentclass[10pt, a4paper, logo, twocolumn, copyright]{deepmind}
\usepackage{kantlipsum, lipsum}
\usepackage{xcolor} 
\usepackage{amsmath, latexsym}
\usepackage{amsfonts}
\usepackage{mathtools}
\usepackage{ntheorem}
\usepackage{dsfont}
\usepackage[dvipsnames]{xcolor}
\usepackage[colorinlistoftodos]{todonotes}
\usepackage{booktabs}
\usepackage{xfrac}
\usepackage{bbm}
\usepackage{bm}
\usepackage{xspace}
\usepackage{ulem}

\usepackage{threeparttable}
\usepackage{array}
\usepackage{caption} 

\usepackage{algorithm}
\usepackage{algorithmicx}

\usepackage[most]{tcolorbox}
\usepackage{xparse}
\usepackage{lipsum}
\usepackage{changepage}
\usepackage{enumitem}

\newcommand{\squishlist}{
   \begin{list}{$\bullet$}
    { \setlength{\itemsep}{0pt}      \setlength{\parsep}{3pt}
      \setlength{\topsep}{3pt}       \setlength{\partopsep}{0pt}
      \setlength{\leftmargin}{1.5em} \setlength{\labelwidth}{1em}
      \setlength{\labelsep}{0.5em} } }

\newcommand{\squishlisttwo}{
   \begin{list}{$\bullet$}
    { \setlength{\itemsep}{0pt}    \setlength{\parsep}{0pt}
      \setlength{\topsep}{0pt}     \setlength{\partopsep}{0pt}
      \setlength{\leftmargin}{2em} \setlength{\labelwidth}{1.5em}
      \setlength{\labelsep}{0.5em} } }

\newcommand{\squishend}{
    \end{list}  }

\newcommand{\two}{(2)}

\newcommand{\ie}{i.e.}
\newcommand{\eg}{e.g.}

\DeclareMathAlphabet{\mathpzc}{OT1}{pzc}{m}{n}

\newcommand{\sequential}{\textit{BP}\xspace}
\newcommand{\sideways}{\textit{Sideways}\xspace}

\newcommand{\simplevgg}{VGG-8\xspace}
\newcommand{\simplecnn}{Simple-CNN\xspace}

\newcommand{\bx}{\bm{x}}

\newcommand{\by}{\bm{y}}

\newcommand{\bh}{\bm{h}}

\newcommand{\bth}{\bm{\theta}}
\newcommand{\timestep}{computation step\xspace}
\newcommand{\timesteps}{computation steps\xspace}
\newcommand{\updatecycle}{update cycle\xspace}
\newcommand{\updatecycles}{update cycles\xspace}
\newcommand{\cyclelength}{cycle length\xspace}

\newcommand{\loss}{\mathcal{L}}
\newcommand{\model}{\mathcal{M}_{\theta}}
\newcommand{\jacobim}{\mathcal{J}}
\newcommand{\activ}{\bm{h}}
\newcommand{\pactiv}{\activ} 

\newcommand{\pseudograd}{\widetilde{\nabla}}

\newcommand{\ndepth}{D}
\newcommand{\pnumber}{p}

\newcommand{\RR}{\mathbb{R}}

\newsavebox\MBox
\newcommand\Cline[2][red]{{\sbox\MBox{$#2$}%
  \rlap{\usebox\MBox}\color{#1}\rule[-1.7\dp\MBox]{\wd\MBox}{2.1pt}}}
\newcommand\ClineY[2][red]{{\sbox\MBox{$#2$}%
  \rlap{\usebox\MBox}\color{#1}\rule[-2.6\dp\MBox]{\wd\MBox}{2.5pt}}}

\definecolor{darkgreen}{rgb}{0.02, 0.4, 0.12}
\definecolor{dandelion}{rgb}{0.62, 0.52, 0.00}

\usepackage[square, numbers]{natbib} 

\graphicspath{{fig/}}

\title{Sideways: Depth-Parallel Training of Video Models}

\correspondingauthor{mateuszm@google.com}


\reportnumber{} 

\author[1]{Mateusz Malinowski}
\author[1]{Grzegorz \'{S}wirszcz}
\author[1]{Jo\~{a}o Carreira}
\author[1]{Viorica P\u{a}tr\u{a}ucean}

\affil[1]{DeepMind}

\begin{abstract}
We propose \sideways, an approximate backpropagation scheme for training video models. In standard backpropagation,  the gradients and activations at every computation step through the model are temporally synchronized. The forward activations need to be stored until the backward pass is executed, 
preventing inter-layer (depth) parallelization. However, can we leverage smooth, redundant input streams such as videos to develop a more efficient training scheme? Here, we explore an alternative to backpropagation; we overwrite network activations whenever new ones, \ie, from new frames, become available. Such a more gradual accumulation of information from both passes breaks the precise correspondence between gradients and activations, leading to theoretically more noisy weight  updates. Counter-intuitively, we show that \sideways training of deep convolutional video networks not only still converges, but can also potentially exhibit better generalization compared to standard synchronized backpropagation.
\end{abstract}

\begin{document}
\maketitle
\balance

\section{Introduction}
The key ingredient of deep learning is stochastic gradient descent (SGD)~\cite{bottou2004large,robbins1951stochastic,zhang2004solving}, which has many variants, including SGD with Momentum~\cite{Sutskever:2013:IIM:3042817.3043064}, Adam~\cite{adam}, and Adagrad~\cite{Duchi:2011:ASM:1953048.2021068}. SGD  approximates gradients using mini-batches sampled from full datasets, and thus it differs from standard gradient descent. Efficiency considerations primarily motivated the development of SGD  as many datasets do not fit in memory.  Moreover, computing full gradients over them would take a long time, compared to mini-batches, \ie, performing SGD steps is often more preferred~\cite{bottou2004large,goyal2017accurate,zhang2004solving}. 
However, SGD is not only more efficient but also produces better models. For instance, giant-sized models trained using SGD are naturally regularized and may generalize better~\cite{he2019rethinking,shen2019object}, and local minima do not seem to be a problem~\cite{choromanska2015loss}. Explaining these phenomena is still an open theoretical problem, but it is clear that SGD is doing more than merely optimizing a given loss function~\cite{zhang2016understanding}. 

In this paper, we propose a further departure from the gradient descent, also motivated by efficiency considerations, which trains models that operate on sequences of video frames. Gradients of neural networks are computed using the backpropagation (BP) algorithm. However, BP operates in a \textit{synchronized} blocking fashion: first, activations for a mini-batch are computed and stored during the forward pass, and next, these activations are re-used to compute Jacobian matrices in the backward pass. Such blocking means that the two passes must be done sequentially, which leads to high latency, low throughput. This is particularly sub-optimal if there are parallel processing resources available, and is particularly prominent if we cannot parallelize across batch or temporal dimensions, e.g., in online learning or with causal models.

\begin{figure}
\centering
\includegraphics[width=0.97\linewidth]{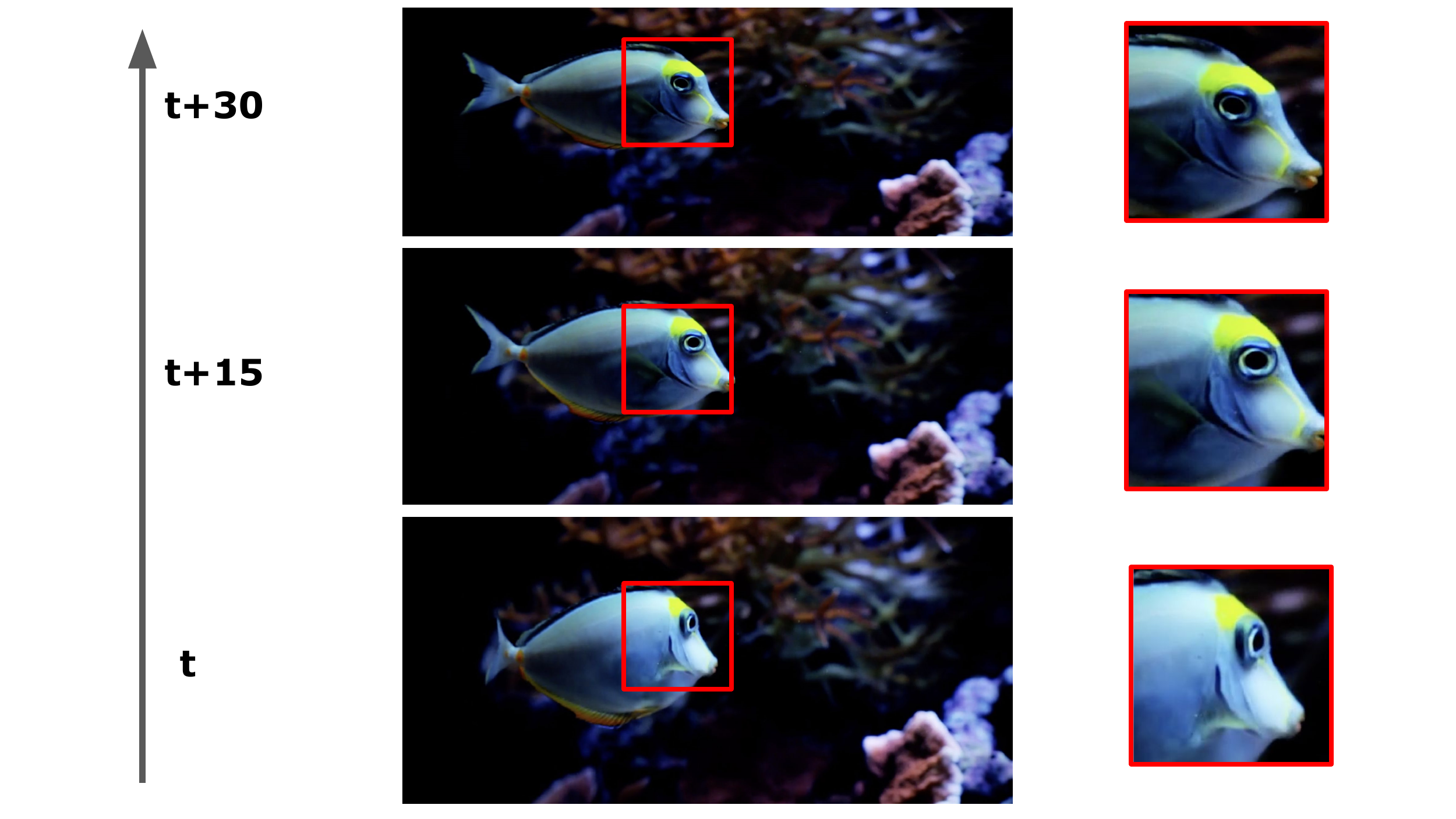}
\caption{Three frames of a fish swimming, sampled 15 frames apart, or about every half a second. Note how little variation there is in the patch within the red square. Can we leverage such redundancies and the smoothness in local neighborhoods of such type of data for more efficient training?  Our results suggest we can and there could be generalization benefits in doing that.
}
\label{fig:teaser}
\end{figure} 

The central hypothesis studied in this paper is whether we can backpropagate gradients based on activations from different timesteps, hence removing the locking between the layers.
Intuitively, one reason this may work is that high frame rate videos are temporally smooth, leading to similar representations of neighboring frames, which is illustrated in~\autoref{fig:teaser}. 

We experiment with two types of tasks that have different requirements in terms of latency: a per-sequence action recognition, and a per-frame autoencoding. In both cases, our models do not use any per-frame blocking during the forward or backward passes. We call the resulting gradient update procedure \sideways, owing to the shape of the data flow, shown in~\autoref{fig:sideways}.

In experiments on action recognition, UCF101~\cite{soomro2012ucf101} and HMDB51~\cite{kuehne2011hmdb}, we have found that training with \sideways not only does not diverge but often has led to improved performance over \sequential models, providing a surprising regularization effect. Such training dynamics create a new line of inquiry into the true nature of the success of SGD, as it shows that it is also not critical to have precise alignment between activations and gradients. Additionally, we show that \sideways provides a nearly linear speedup in training with depth parallelism on multiple GPUs compared to a \sequential model using the same resources. We believe that this result also opens up possibilities for training models at higher frame rates in online settings, e.g., where parallelization across mini-batches is not an option.

We use per-frame autoencoding task to investigate the effect of the blocking mechanism of \sequential models in tasks where the input stream cannot be buffered or where we require immediate responses. This is particularly problematic for \sequential if the input stream is quickly evolving, \ie, the input change rate is higher than the time required to process the per-step input. In this case, the blocking mechanism of \sequential will result in discarding the new inputs received while the model is being blocked processing the previous input.  However, this is considerably less problematic in \sideways due to its lock-free mechanism. We run experiments on synthetically generated videos from the CATER dataset~\cite{girdhar2019cater}, where we observe that \sideways outperforms the \sequential baseline.

\section{Related Work}

Our work connects with different strands of research around backpropagation, parallelization and video modelling. We list here a few of the most relevant examples.  

\vspace{2mm}
\noindent \textbf{Alternatives to backpropagation.}
Prior work has shown that various modifications of the `mathematically correct' backpropagation can actually lead to satisfactory training. For instance, some relaxations of  backpropagation implemented with a fixed random matrix yield a surprisingly good performance on MNIST~\cite{lilicrap}.
There is also a recent growing interest in building more biologically-plausible or model-parallel approaches to train networks. This includes Feedback Alignment~\cite{lilicrap}, Direct Feedback Alignment~\cite{NIPS2016_6441}, Target Propagation~\cite{bengio2015towards}, Kickback~\cite{balduzzi2014kickback}, Online AM~\cite{choromanska2018beyond}, Features Replay~\cite{huo2018training}, Decoupled Features Replay~\cite{belilovsky2019decoupled}, and Synthetic Gradients~\cite{Jaderberg:2017:DNI:3305381.3305549},  where various decouplings between forward or backward pass are proposed. A good comparative overview of those frameworks is presented in~\cite{Czarnecki2017UnderstandingSG}. Another recent innovative idea is to meta-learn local rules for gradient updates~\cite{metz2018learning}, or to use either self-supervised techniques~\cite{oord2018representation} or local losses to perform gradient-isolated updates locally~\cite{lowe2019putting,nokland2019training}.
Asynchronous distributed SGD approaches like Hogwild \cite{recht2011hogwild} also do not strictly fit into clean backprop as they allow multiple workers to partially overwrite each others weight updates, but provide some theoretical guarantees as long as these overwrites are sparse. However, most of these prior works are applied to visually simpler domains, some require buffering activations over many training steps, or investigate local communication only. In contrast, here, we take advantage of the smoothness of temporal data.  Moreover, we investigate a global, top-down, and yet asynchronous communication between the layers of a neural network during its training without buffering activations over longer period and without auxiliary networks or losses. This view is consistent with some mathematical models of cortex~\cite{betti2019backprop,kubilius2018cornet,larkum2013cellular,tomita1999top}.  We also address forward and backward locking for temporal models. Finally, most of  the works above can also potentially be used together with our \sideways training, which we leave as a possible future direction.

\vspace{2mm}
\noindent \textbf{Large models.} Parallelism has grown in importance due to the success of gigantic neural networks with billions of parameters~\cite{vaswani2017attention}, potentially having high-resolution inputs~\cite{real2019regularized}, that cannot fit into individual GPUs. Approaches such as GPipe~\cite{DBLP:journals/corr/abs-1811-06965} or DDG~\cite{conf/icml/HuoGYH18} show that efficient pipelining strategies can be used to decouple the forward and backward passes by buffering activations at different layers, which then enables the parallel execution of different layers of the network. Similarly, multiple modules of the network can be processed simultaneously on activations belonging to different mini-batches~\cite{conf/icml/HuoGYH18}. Such pipelining reduces the training time for image models but at the cost of increased memory footprint.

\vspace{2mm}
\noindent \textbf{Efficient video processing.} Conditional computation~\cite{bengio2015conditional} or hard-attention approaches can increase efficiency~\cite{malinowski2018learning,mnih2014recurrent} when dealing with large data streams. These are, however, generic approaches that do not exploit the temporal smoothness of sequential data such as video clips~\cite{wiskott2002slow}. For video, sampling key frames is shown to be a quite powerful mechanism when performing  classification~\cite{korbar2019scsampler,DBLP:journals/corr/abs-1907-13369}, but may not be appropriate if a more detailed temporal representation of the input sequence is needed~\cite{girdhar2019cater}. Recently, a deep decoupled video model~\cite{eccv2018massively} has been proposed that achieves high throughput and speed at inference time, while preserving the accuracy of sequential models. However,~\cite{eccv2018massively} uses regular backprop, and hence does not benefit parallelization fully, \ie, backprop still blocks the computations, and requires buffering activations during the forward pass. In this paper, we build  upon~\cite{eccv2018massively} that use parallel inference, but go further and make both inference and learning depth-parallel. Note that, if we only consider inference, \sideways reduces to~\cite{eccv2018massively}.

\section{Sideways}
In this section, we define the formulation of our problem and formalize both algorithms: \sequential and \sideways. 
\subsection{Notation and Definitions}
\newcommand{\modulefun}{H}

We consider the following general setting:
\begin{itemize}
\item  a finite input time-series $\bx = (\bx^t)_{t=1}^K, \bx^t \in \RR^d$, \eg, a video clip with $d = \text{height} \times \text{width} \times 3$,
\item a finite output time-series $\by = (\by^t)_{t=1}^K, \by^t \in \RR^{d_y}$, e.g., an action label; in the action recognition task, in our work, we use the same label over the whole video clip, \ie, $\by^t = \by^{t+1}$ for all $t$,
\item  a frame-based neural network $\model : \RR^d \rightarrow \RR^{d_y}$ that transforms the input signal $\bx^t$ into logits $\bh_D^t = \model(\bx^t)$, and is defined by a composition of modules 
\begin{equation*}
    \model(\bx^t) = \modulefun_D(\cdot, \theta_\ndepth) \circ \modulefun_{\ndepth-1}(\cdot, \theta_{\ndepth-1}) \circ \ldots \circ \modulefun_1(\bx^t, \theta_1) 
\end{equation*}
where:
\begin{itemize}
    \item each module, or layer, $\modulefun_i(\cdot, \cdot)$ is a function $\modulefun_i : \RR^{d_{i-1}} \times \RR^{\pnumber_{i}} \rightarrow \RR^{d_{i}}$, $i=1, \ldots \ndepth$,
    \item $\theta_i \in \RR^{\pnumber_i}$, $i=1,\ldots,\ndepth$ are the (trainable) parameters, and we use $\theta$ for all the parameters,
    \item $\circ$ is a composition, \ie, $G \circ F(\bx) = G(F(\bx))$
\end{itemize}
and
\item  a loss function $\loss : \RR^{d_y} \times \RR^{d_y} \rightarrow \RR$, \eg,  $\loss(\bh, \by) = ||\bh - \by||^2$, or $\loss(\bh, \by) = -\sum_i p((\bh)_i)\log q(\by_i)$.
\end{itemize}
We extend the notation above to
$\activ_i^t = \modulefun_i(\cdot, \theta_i) \circ \modulefun_{i-1}(\cdot, \theta_{i-1}) \circ \ldots \circ \modulefun_1(\bx^t, \theta_1)$.
 
To avoid the common confusion coming from using the same letters to denote both the function formal arguments and actual values of the variables, 
we will use bold font for the latter, \eg, $x$ to denote a formal argument and $\bx$ for its actual value.
We also use the following notation for the derivatives of the functions $\modulefun_i$. Let 
$\jacobim_h H(\bh, \bth) = \left. \frac{\partial \modulefun(h, \bth)}{\partial h} \right|_{h = \bh}$ be the Jacobian matrix of $H(h, \theta)$ with respect to the variable $h$ evaluated at $h=\bh$, $\theta = \bth$. Similarly, $\jacobim_{\theta} H(\bh, \bth) = \left. \frac{\partial \modulefun(\bh, \theta)}{\partial \theta} \right|_{\theta = \bth}$  denote the Jacobian matrix of $H(h, \theta)$ with respect to the variable $\theta$ evaluated at $h=\bh$, $\theta = \bth$. We will use the same notation for the gradient $\nabla$.
 
Finally, to train neural networks, we base our computations on the empirical risk minimization framework, \ie\, $\mathcal{R}(\model) = E_{x,y}[\loss(\mathcal{M}_{\theta}(x), y)] \approx \sum_{\bx, \by \sim \mathcal{D}}\frac{1}{K}\sum_{t=1}^K\loss(\bh_D^t, \by^t)$,
where $\mathcal{D}$ is a training set.

\subsection{Update Cycle}
For simplicity, we assume in our modelling a constant time for a layer (or some set of layers organized into a module) to fully process its inputs, both in the forward or backward pass and call this a \textit{\timestep}.
We define the \textit{\updatecycle} as the sequence of \timesteps that a given data frame is used to update all the layers, and the \textit{\cyclelength} as the number of \timesteps in the \updatecycle.   Hence, the \cyclelength depends only on the depth of the network $D$ and is equal to $2D-1$ \timesteps. \autoref{fig:sideways} illustrates a single \updatecycle with nine \timesteps for both models.

\subsection{The \sequential algorithm (`regular' backpropagation) }
The \sequential algorithm refers to regular training of neural networks. Here, due to the synchronization between the passes, computations are blocked each time a data frame is processed. This is illustrated in \autoref{fig:sideways} (left). Whenever the first frame is processed, here indicated by the blue square, the computations are blocked in both forward and backward passes over the whole \updatecycle.

With our notation, the standard backpropagation formula becomes
\begin{eqnarray*}
\label{eq::backprop}
\nabla_{\theta_i}^t \loss &=& \nabla_{\theta_i} \loss(\model(\bx^t), \by^t) |_{\theta = \bth } =\\
&& \nabla_{h_D} \loss(\activ_\ndepth^t, \by^t) \cdot  \jacobim_{h_{\ndepth-1}} \modulefun_\ndepth(\activ_{\ndepth-1}^t, \bth_\ndepth) \cdot \\
&& \jacobim_{h_{\ndepth-2}} \modulefun_{\ndepth-1}(\activ_{\ndepth-2}^t, \bth_{\ndepth-1}) \cdot \\
&& \vdots\\
&& \jacobim_{h_i} \modulefun_{i+1}(\activ_i^t, \bth_{i+1}) \cdot \\
&& \jacobim_{\theta_i} \modulefun_{i}(\activ_{i-1}^t, \bth_{i})
\end{eqnarray*}
with the update rule $\theta_i := \theta_i - \alpha \frac{1}{K}\sum_{t=1}^K\nabla_{\theta_i}^t \loss$,
where $\alpha$ is the learning rate, and $K$ is the length of the input sequence.

We can compactly describe the algorithm above with the following recursive rules
 \begin{eqnarray}
  \label{eq:sequential_backprop_weight}
 \nabla_{\theta_i}^t \loss &=& \nabla_{h_{i}}^t \loss \cdot \jacobim_{\theta_i} \modulefun_{i}(\activ_{i-1}^{t}, \bth_{i})\\
  \label{eq:sequential_backprop_backpass}
 \nabla_{h_{i-1}}^t \loss &=& \nabla_{h_{i}}^t \loss \cdot \jacobim_{h_{i-1}} \modulefun_{i}(\activ_{i-1}^{t}, \bth_{i})
 \end{eqnarray}
 where $\bh^t_0 = \bx^t$.
 However, we do not compute Jacobian matrices explicitly; instead efficient vector matrix multiplications are used to backpropagate errors from the loss layer towards the input~\cite{abadi2016tensorflow}.
 
\subsection{\sideways algorithm}
\begin{figure*}[t]
\begin{center}
\begin{tabular}{c@{\ }c@{\ }c}
\hspace{-10mm}\includegraphics[width=0.42\linewidth]{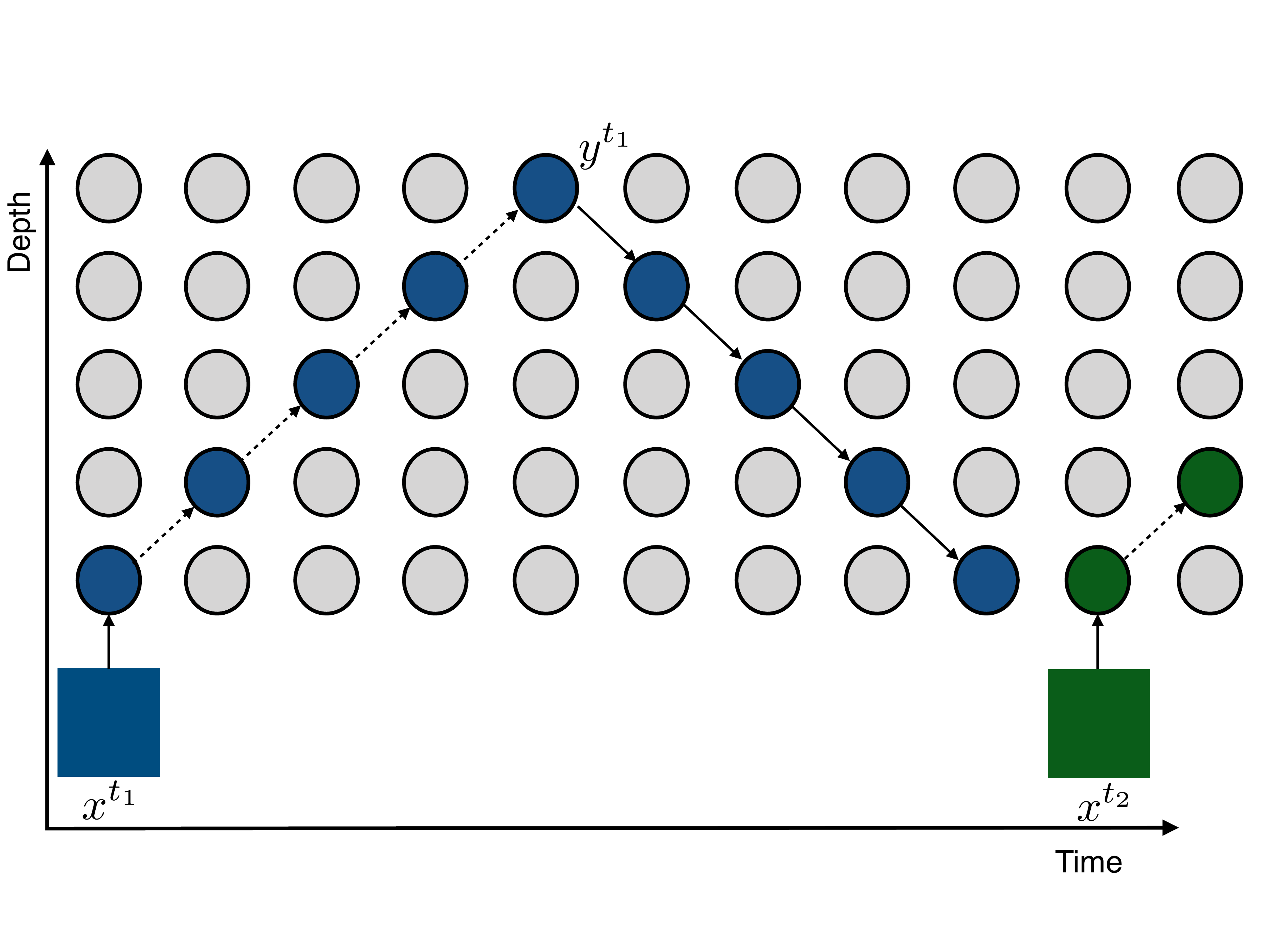} & \hspace{8mm}
\includegraphics[width=0.42\linewidth]{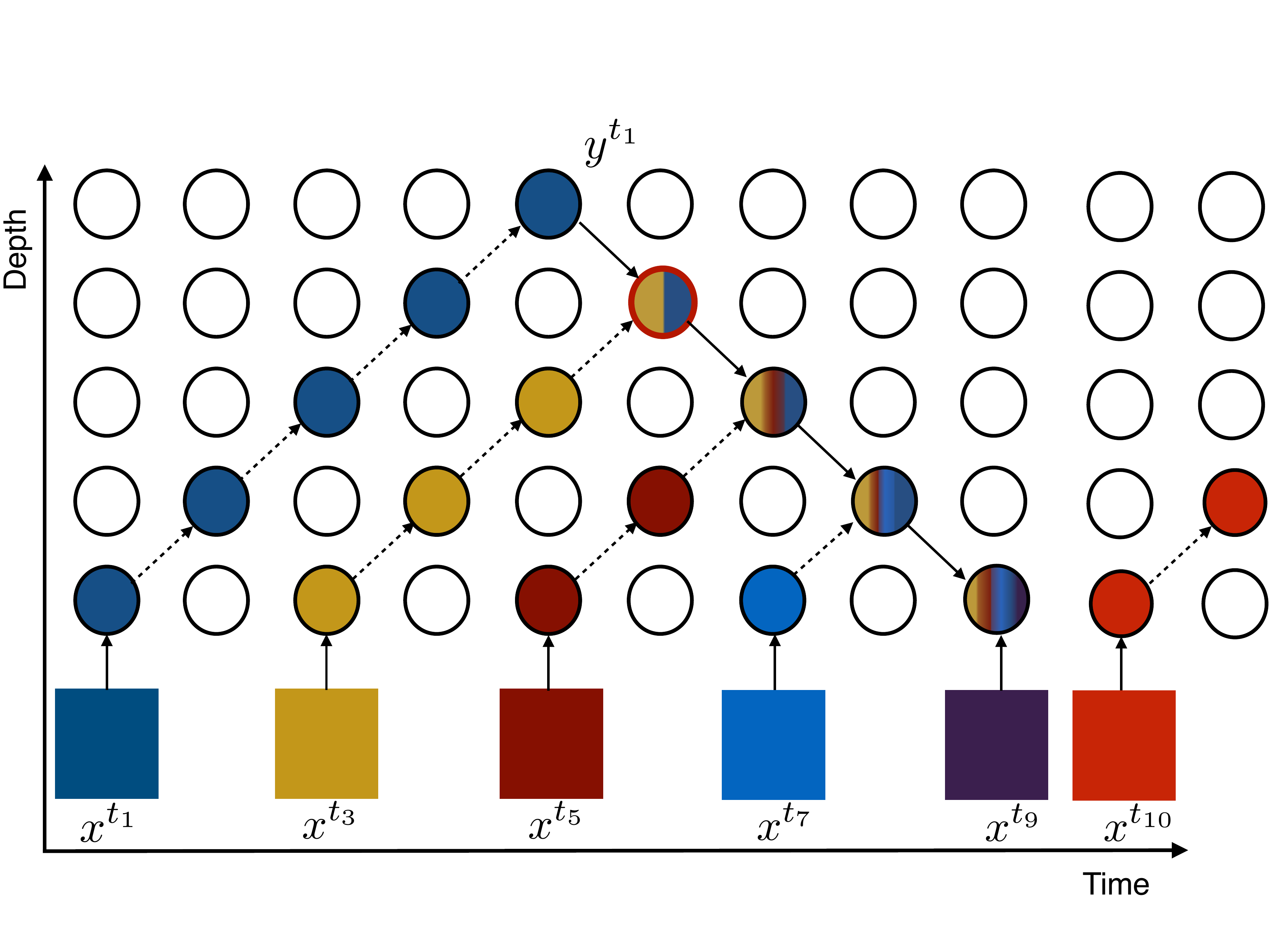}
\end{tabular}
\end{center}
\vspace{-7mm}
\caption{
From left to right. Standard (\sequential) and fully pipelined (\sideways) approaches to temporal training and inference. We show a single \updatecycle, and the beginning of the next cycle.
Both architectures are unrolled in time. Colorful squares indicate data frames. Circles indicate `regular' or \sideways modules. Dotted arrows show how information is passed between layers and time steps in forward pass. Solid arrows show the same in backward pass. In \sideways (right), we only exemplify a single update path with the arrows, and use empty circles for all other units. Gray circles denote blocked modules, \ie, units waiting for forward or backward pass. 
Note that for \sequential, we use the same color for all the units on the data path, in both the forward and the backward passes, to highlight that all the layers work on information originated in a single data frame, the blue one. Differently, the back-pass in \sideways shows circles with many colors to illustrate that information from different data frames is combined in one update cycle. 
For instance, combining `blue gradient' with `yellow activations' yields `blue-yellow gradient' (6th \timestep and second-last unit).
Best viewed in color. 
}
\label{fig:sideways}
\end{figure*}

We aim at pipelining computations for the whole \updatecycle during training and inference. \sideways removes synchronization by continuously processing information, either in the forward or backward pass. This is illustrated in \autoref{fig:sideways} (right). Once a data frame is available, it is immediately processed and sent to the next layer, `freeing' the current layer so it can process the next data frame. Hence, in the first \timestep of the \updatecycle, a data frame $\bx^t$ is processed by the first \sideways module, freeing resources and `sending' $\bh^t_1$ to the second \sideways module at \timestep $t+1$.
At \timestep $t+1$, the first module can now take the next data frame $\bx^{t+1}$ for processing, and, simultaneously, the second module processes $\bh_1^t$; this step results in two representations $\bh_2^t$ and $\bh_1^{t+1}$. Please note that our notation $\bh_2^t$ does not indicate the current \timestep but instead that the representation has originated at $\bx^t$. We continue the same process further during the training. This is illustrated in \autoref{fig:sideways}, where we use color-encoding to track where the information being processed has originated from. Dotted arrows represents the forward pass.

For simplicity, we assume that the computation of the loss takes no time and does not require an extra computation cycle. In such setting the activation arriving at the loss function computing module at timestep $t$ is $\bh_D^{t-\ndepth+1}$, an activation spawned by the frame $\bx^{t-\ndepth+1}$. Once this final representation $\bh_D^{t-\ndepth+1}$ is computed at \timestep $t$, we calculate its `correct' gradient $\nabla_{h_\ndepth}^t\loss(\bh_D^{t-\ndepth+1}, \by^{t})$, and we backpropagate this information down towards the lower layers of the neural network. This computational process is illustrated in \autoref{fig:sideways} (right) by the solid arrows.

Let us formalize this algorithm in a similar manner to the `regular' backpropagation.
In the \sideways algorithm the gradient $\nabla_{\theta_i} \loss |_{(\bx^t, \bth_i)}$ is replaced with a {\it pseudo-gradient} $\pseudograd_{\theta_i} \loss |_{(\bx^t, \bth_i)} $, defined as follows
\begin{eqnarray*}
\hspace{-1mm}
\label{eq::sideways_backprop}
\pseudograd_{\theta_i}^t \loss &=& 
\hspace{-1mm}
\nabla_{h_D} \loss(\pactiv_\ndepth^{t_i - \ndepth + 1}, \by^{t_i}) \cdot  \jacobim_{h_{\ndepth-1}} \modulefun_\ndepth(\pactiv_{\ndepth-1}^{t_i-\ndepth+1}, \bth_\ndepth) \cdot \\
&& \jacobim_{h_{\ndepth-2}} \modulefun_{\ndepth-1}(\pactiv_{\ndepth-2}^{t_i-\ndepth+3}, \bth_{\ndepth-1}) \cdot \\
&& \vdots\\
&& \jacobim_{h_i} \modulefun_{i+1}(\pactiv_i^{t-i-1}, \bth_{i+1}) \cdot \\
&& \jacobim_{\theta_i} \modulefun_{i}(\pactiv_{i-1}^{t-i+1}, \bth_{i})
\end{eqnarray*}
where $t_i = t + i - \ndepth$.

The equations above can next be written succinctly and recursively as the \sideways backpropagation rules
 \begin{eqnarray}
 \label{eq:sideways_backprop_weight}
\pseudograd_{\theta_i}^t \loss &=& \Cline[blue]{\pseudograd_{h_{i}}^{t-1} \loss} \cdot \ClineY[yellow]{\jacobim_{\theta_i} \modulefun_{i}(\pactiv_{i-1}^{t-i+1},\bth_{i})}\\
 \label{eq:sideways_backprop_backpass}
  \pseudograd_{h_{i-1}}^t \loss &=& \Cline[blue]{\pseudograd_{h_{i}}^{t-1} \loss} \cdot \ClineY[yellow]{\jacobim_{h_{i-1}} \modulefun_{i}(\pactiv_{i-1}^{t-i+1}, \bth_{i})}
 \end{eqnarray}
where $\pseudograd^{t-1}_{h_D}\loss = \nabla_{h_D}\loss(\bh_D^{t-\ndepth+1}, \by^t)$, and $\bh^t_0 = \bx^t$.

In the equations above, we use a color-encoding similar to \autoref{fig:sideways} (right) to indicate that we combine information originated from different time steps. For instance, information originated in `blue' and `yellow' input frames is combined (6-th \timestep and second-last unit) as indicated by the red circle in \autoref{fig:sideways} (right)). By following the arrows we can track the origins of the combined information. 

Due to the nature of these computations, we do not compute proper gradients as the \sequential algorithm does, but instead we compute their more noisy versions, $\pseudograd_{\bh_i}\mathcal{L} = \nabla_{\bh_i}\mathcal{L} + \epsilon_{i}(\bx)$, which we call pseudo-gradients. The amount of noise varies with respect to the smoothness of the input $\bx$, and the number of the layer  $i$. That is, deeper layers have less noisy pseudo-gradients, and \eg, the pseudo-gradient of the final layer is exact.

We organize training as a sequence of episodes. Each episode consists of one or more \updatecycles, runs over the whole sampled video clip $\bx$ or its subsequence, and ends with the weights update. We assume the input $\bx$ is smooth within the episode, \eg, $\bx$ is a video of an action being performed with a reasonable frame-rate. We `restart' \sideways by setting up all the activations and pseudo-gradients to zero whenever we sample a new video to avoid aliasing with a pseudo-gradient originated from a data frame from another video clip, and thus breaking our assumptions about the smoothness of the input sequence.  Mini-batching can optionally be applied in the usual way.

We average gradients computed at each layer over all \timesteps within the episode, \ie, 
\begin{align}
    \label{eq:pseudogradient_average}
    \pseudograd_{\theta_i}\loss = \frac{1}{L}\sum_{t=1}^L\pseudograd_{\theta_i}^t\loss
\end{align}
where $L$ is the length of the episode. In our experiments we consider two cases. In the \textit{classification} task, the episode is the same as the sampled sequence, \ie, $L=K$. In the \textit{auto-encoding} task, the episode is a single data frame, \ie, $L=1$.
We use pseudo-gradients $\pseudograd_{\theta_i}\mathcal{L}$ for the weight updates, \ie, $\theta_i  := \theta_i - \alpha \pseudograd_{\theta_i}\loss$.

\begin{figure*}[t]
\begin{center}
\begin{tabular}{c@{\ }c@{\ }c}
\hspace{-8mm}\includegraphics[width=0.42\linewidth]{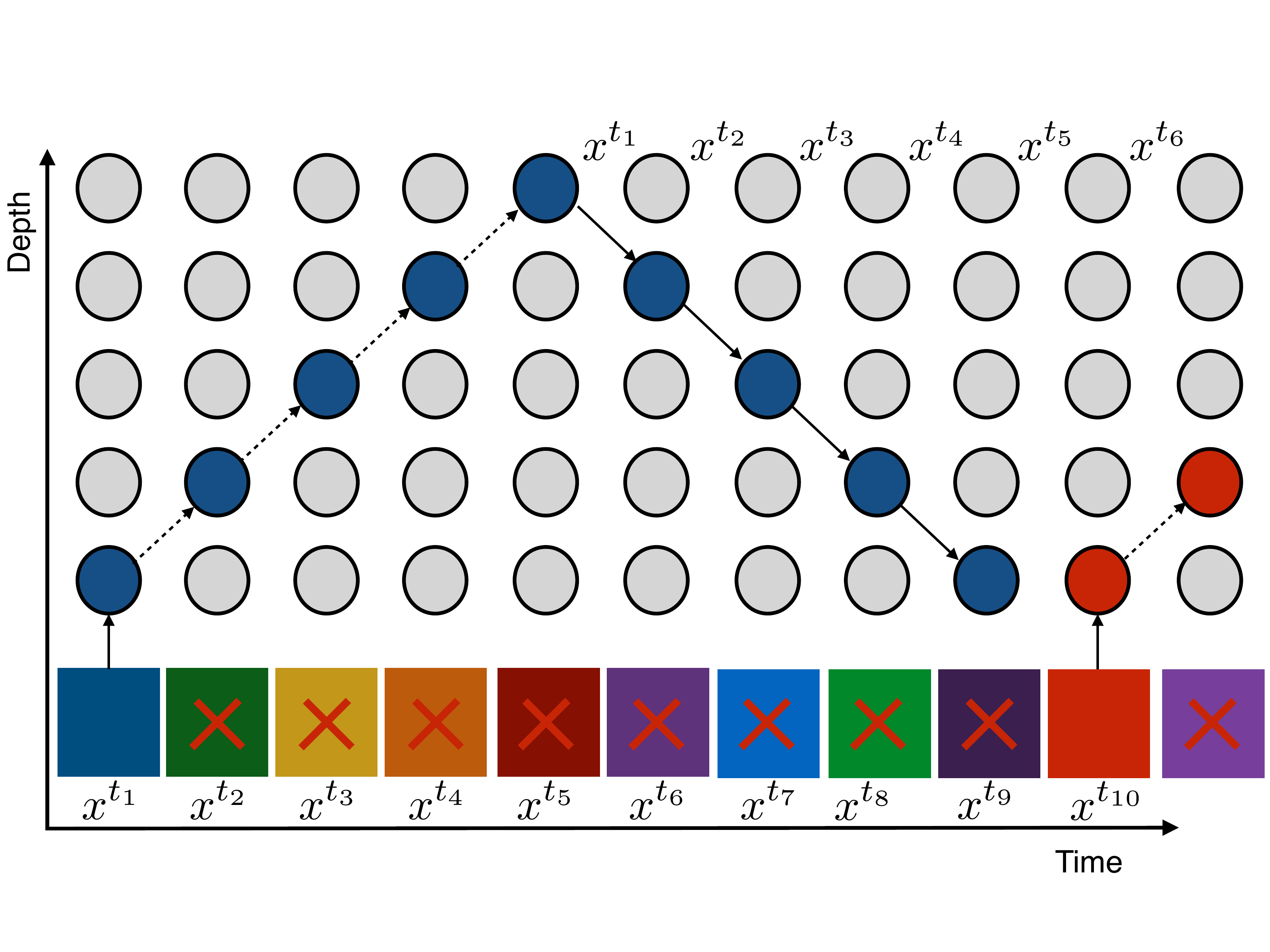} & \hspace{8mm}
\includegraphics[width=0.42\linewidth]{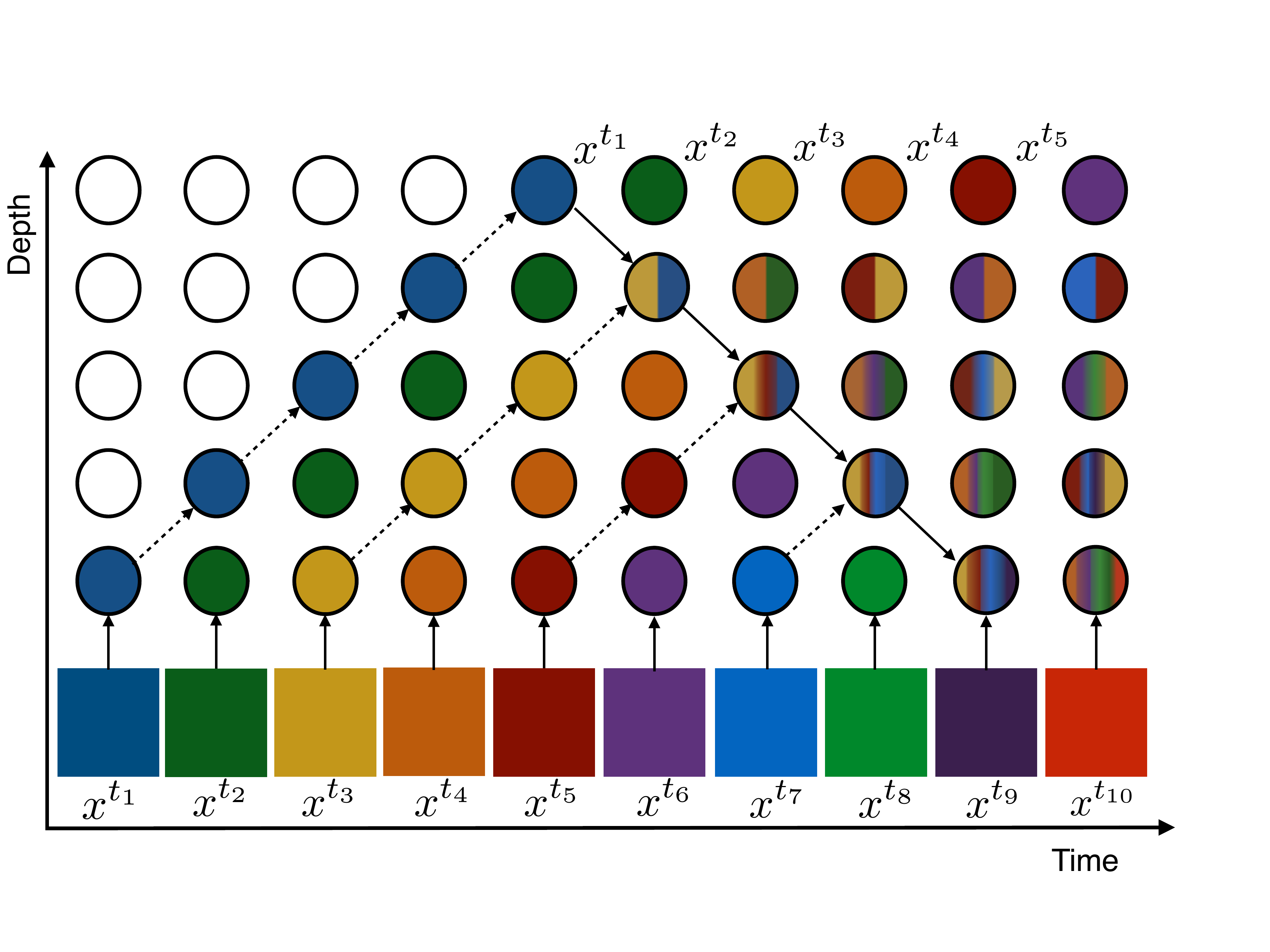}
\end{tabular}
\end{center}
\vspace{-7mm}
\caption{
From left to right. \sequential and \sideways approaches to temporal training and inference. In the figure, we illustrate the auto-encoding task, where the network needs to synthesize input frames.  
Crossed frames denote input data ignored because the system cannot operate in real-time to process all the inputs (left).
In contrast, \sideways works simultaneously at full capacity once the pipeline is full; and since we show the beginning of the episode some units are unused (empty circles) due to the shape of the data flow (right).
All the colors and arrows have the same meaning as in \autoref{fig:sideways}. 
Best viewed in color.
}
\label{fig:sideways_vs_sequential_evolve_input}
\end{figure*}

\autoref{fig:sideways_vs_sequential_evolve_input} (right) illustrates the situation when the pipeline is full and suggests, the information flow is tilted sideways. Therefore, there is no information available in the upper layers at the beginning of the sequence (empty circles in the figure). For that reason, we modify \autoref{eq:pseudogradient_average} by including a binary mask, \ie, $\pseudograd_{\theta_i}\loss = \frac{1}{\gamma_i}\sum_{t=1}^L\gamma_i^t\pseudograd_{\theta_i}^t\mathcal{L}$, where $\gamma_i = \sum_t \gamma_i^t$. The mask is zero for unavailable gradients. For similar reasons, to avoid gradient computations whenever suitable information is unavailable, we modify \autoref{eq:sideways_backprop_backpass} with 
$\pseudograd_{\bh_i}^t\mathcal{L}  = \gamma_i^t\pseudograd_{h_{i}}^{t-1} \loss \cdot \jacobim_{h_{i-1}} \modulefun_{i}(\pactiv_{i-1}^{t-i+1}, \bth_{i})$. Without masking, we have  observed more unstable training in practice.
\vspace{2mm}
\newline\noindent
\textbf{Intuitions.} As we make the input sequence increasingly more smooth, in the limits, each data frame has identical content. In such a case, since $\epsilon_i(\bx)=0$, pseudo-gradients equal gradients, and our algorithm is the same as the `regular' backpropagation. In practice, if the input sequence has different data frames, we assume that two consecutive frames are similar, and especially essential features are slowly evolving, sharing their semantics within the neighborhood~\cite{wiskott2002slow}.


\section{Experiments}
We investigate both algorithms – \sequential and
\sideways – on several benchmarks. Since, to the best of our knowledge, this is possibly the first work on depth-parallel training on challenging video tasks, we focus on simple convolutional networks, and aim to explore the training dynamics instead of seeking state-of-the-art results. We leave data augmentation, additional features such as optical flow, or pre-training on large datasets~\cite{carreira2017quo, diba2019dynamonet, jing2018self, karpathy2014large, simonyan2014two} for future work. We compare frame-based video models~\cite{jing2018self,karpathy2014large,simonyan2014two} trained either using \sequential or \sideways that are \textit{trained from scratch} and using standard setups.

\subsection{Task}
We benchmark our algorithms on two different tasks and three datasets.
\vspace{2mm}
\newline
\noindent
\textbf{Classification.}
We start with the classical classification task, here, in the form of action recognition. Since the classification is at the core of regular supervised learning, we believe, any alternative, sequential or parallel, to SGD should be evaluated on this common task. \autoref{fig:sideways} illustrates both algorithms under the classification scenario. Differently to the next, auto-encoding task, here, we test the networks under the regular circumstances, where each frame is always guaranteed to be processed by the neural network.
\vspace{2mm}
\newline
\noindent
\textbf{Auto-encoding.}
While majority of our key results are on the classification task, it is also informative to validate \sideways on tasks where the target output is continuously changing with the input.
As a proof of concept, we experiment here with the simpler task of auto-encoding. To clearly illustrate advantages of \sideways training, and for the sake of simplicity, we assume that the input frame rate and the processing time for each individual neural network layer are equal. This is shown in~\autoref{fig:sideways_vs_sequential_evolve_input}.
If the stream is a sequence $(x^{t_1}, x^{t_2},  x^{t_3}, \ldots)$, $D$ is the number of modules, then \sequential blocks the processing of the input for $2 (D - 1)$ \timesteps, hence ignoring data frames between $t_1$ and $t_{10}$ during training for $D=5$. This is illustrated in~\autoref{fig:sideways_vs_sequential_evolve_input} (left). In contrast, \sideways pipelines computations and uses all the data frames in both training and inference modes. This often results in superior performance of \sideways under the circumstances mentioned above. Finally, by comparing Figures~\ref{fig:sideways} and~\ref{fig:sideways_vs_sequential_evolve_input}, we can clearly see the \sideways algorithm behaves identically, even if we artificially introduce the blocking mechanism described above.

\subsection{Datasets}
We choose to benchmark our models of computations on the following video datasets. On one hand, first two datasets have enough complexity and realism. On the other hand, we can easily train frame-based video models on all the following datasets.
\vspace{2mm}
\newline
\noindent
\textbf{HMDB51} is a widely used dataset for action recognition that has 6770 video clips representing 51 actions~\cite{kuehne2011hmdb}. Video clips run at $30$fps. In our experiments, we use the first train and test splits, similar to ablation studies in~\cite{jing2018self,simonyan2014two}. This is the smallest real-world dataset in our experiments, and since training is fast, we therefore  use mainly this setting to study our models in details.
\vspace{2mm}
\newline
\textbf{UCF101} is another popular dataset for action recognition~\cite{soomro2012ucf101}. It has 13320 videos clips and 101 human actions. Actions include pizza tossing, typing, playing cello, skiing, etc. Default frame-rate is $25$fps, with the duration of $7.21$sec on average. Each frame has resolution $320$-by-$240$. We use train and test splits in our studies. In our experiments, we find this dataset to be of particular interest due to its size, complexity, and realism.
\vspace{2mm}
\newline
\textbf{CATER} dataset~\cite{girdhar2019cater} provides synthetically generated videos of moving simple 3D objects. We use only the video frames and we set up an unsupervised auto-encoding task. These videos have two desired properties -- i) they are visually simple, and ii) they have diverse motion patterns of various objects -- making it an excellent benchmark for \sideways. We use the pre-generated video sequences provided by the authors in the \textit{all\_actions} subset, consisting of 3861 training video sequences and 1639 test sequences, each with 300 frames, having $320$-by-$240$ pixels resolution at $24$fps. 

\subsection{Architectures}
\label{sec::architectures}

For the purposes of our study we have experimented with two standard convolutional network architectures.
The first one is organized into 6 \sideways modules, another one with 8 \sideways modules. Note, however, that we can use more than one trainable layers inside a single \sideways module.

\vspace{2mm}
\noindent
\textbf{\simplecnn} is a simple and fast baseline consisting of $5$ convolutional layers with kernel size $3x3$ followed by global average pooling and a softmax on the linear layer that projects the internal representation into classes. The convolutional layers have the following number of channels: $(32, 64, 64, 128, 256)$. To reduce resolution progressively in the network, we use striding $2$ in every second layer starting from the first one. 

For the auto-encoding experiments, we train a simple encoder-decoder architecture having the same five convolutional blocks followed by symmetrical five deconvolutional blocks. We use \sideways blocks only for the convolutional encoder; the decoder layers are connected all in a single sequential block, and hence the decoder-block is trained with a regular \sequential with `correct' gradients. For simplicity, we also assume the whole decoder takes just a single \timestep. We use this setting to better investigate the quality of the features extracted by the \sideways encoder.
\vspace{2mm}
\newline
\textbf{VGG-net} refers to \simplevgg, which is a direct re-implementation of the RGB network in the original two-stream model~\cite{simonyan2014very} with the addition of batchnorm in every VGG-block (in between the convolution and the ReLU). 

\vspace{2mm}
\noindent \textbf{Implementation details.} By default, we use gradient clipping by value $1.0$. We use a bias term only on the last linear layer used to produce logits. We considered three different popular optimizers: SGD, SGD with momentum, and Adam~\cite{adam}. We use default hyperparameters for all of them and our observation is that \sideways is stable under all the considered training algorithms so we report results with Adam only. We use a warm-up scheme by linearly interpolating learning rate in each iteration between $0$ at the beginning and the initial value at the $5$-th epoch. Later we drop the learning rate by dividing it by $10$ at $100$-th and $200$-th epochs. We use Tensorflow 2~\cite{abadi2016tensorflow} to implement our framework.

In some experiments we use batchnorm~\cite{ioffe2015batch} and dropout~\cite{srivastava2014dropout} in linear layers ($0.9$ as~\cite{simonyan2014two}). In most training experiments, we use a single GPU (V100) -- but measure parallelization speedup later in the section using multiple GPUs. We decouple the weight decay term from the loss~\cite{loshchilov2018decoupled} and find that some amount of weight decay is beneficial for generalization of the VGG-net models. We do a hyper-parameter search over the weight decay coefficient (possible values are: $0.0, 10^{-4}, 10^{-3}, 10^{-2}$), and over the initial learning rate (either $10^{-4}$ or $10^{-5}$), fairly for both types of training. 

We train on videos with the per-frame resolution $112$-by-$112$ and a sequence length $64$ randomly cropped from the whole video clip as a contiguous subsequence. We also use random flipping during training, and this augmentation is performed consistently over all the frames in every clip. At test time we simply take a center crop of a video frame. In all cases, we use torus-padding (video frames are repeated) whenever a sampled video clip is shorter than $64$ and at test time we evaluate over the whole video sequence. In practice we use a fixed batch size of 8 videos for training.  

For the auto-encoding experiments with CATER, we use square crops of 240$\times$240 pixels, extracted randomly during training and central crops of the same size during testing.

\subsection{Results (Classification)}
We have evaluated networks trained with \sideways and \sequential according to the regular accuracy metric. Moreover, to gain a better understanding, and to show how general the \sideways training is, we have conducted  several different experiments measuring not only accuracy but also training dynamics and robustness of the method. 
\begin{table}[t]
    \centering
    \resizebox{0.99\linewidth}{!}{%
    \begin{tabular}{lll}
        \toprule
        HMDB51 & \sequential & \sideways \\
        \midrule
        \simplecnn & 17.2  &  16.5 \\
        \simplevgg & 24.6 & 25.8 \\
        \midrule
        3DResNet (scratch)~\cite{hara2018can,jing2018self}  & 17.0 & - \\
        \toprule
        \toprule
        UCF101 & \sequential & \sideways \\
        \midrule
        \simplecnn & 40.7 & 42.16 \\
        \simplevgg & 49.1 & 53.8 \\
        \simplevgg + Dropout (0.9) & 56.0 & 58.2 \\
        \midrule
        VGG-8 (scratch)+ Dropout (0.9)~\cite{simonyan2014two}   & 52.3 & - \\
        3DResNet (scratch)~\cite{hara2018can,jing2018self} & 42.5 & - \\
        \bottomrule
    \end{tabular}%
    }
    \caption{
    Comparison of our implementation of two architectures using \sideways and \sequential computational models on different datasets. For reference, we also report similar models from  prior work~\cite{hara2018can,jing2018self,simonyan2014two}. We report accuracy in $\%$.
    }
    \label{tab:sequential_vs_sideways}
\end{table}

\begin{figure*}[t]
\begin{center}
\hspace{-6mm}
\scalebox{0.85}{
\begin{tabular}{l@{\ }l@{\ }l@{\ }c@{\ }c@{\ }c@{\ }c}
 \toprule
 & \includegraphics[width=0.26\linewidth]{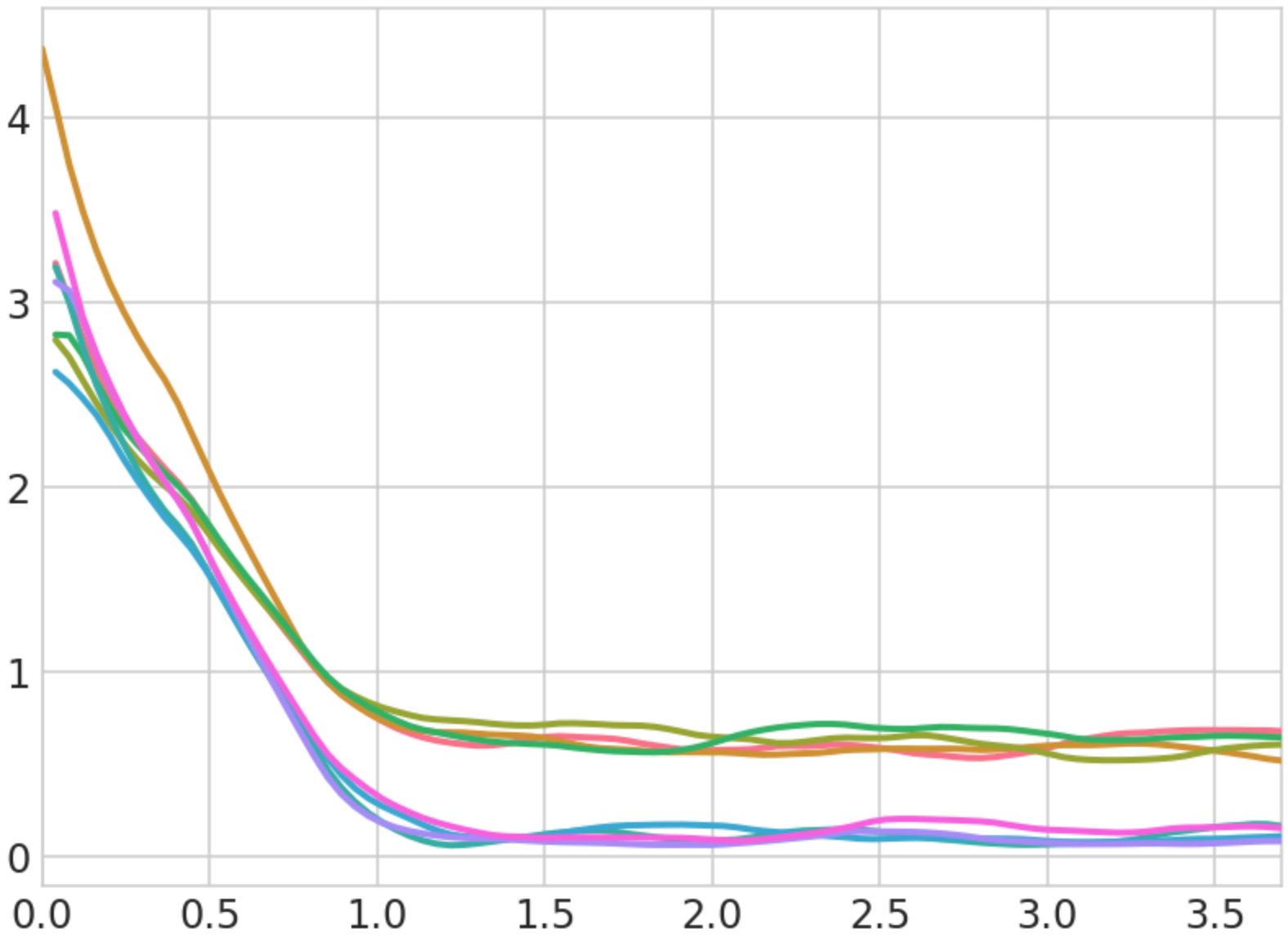} &
\includegraphics[width=0.26\linewidth]{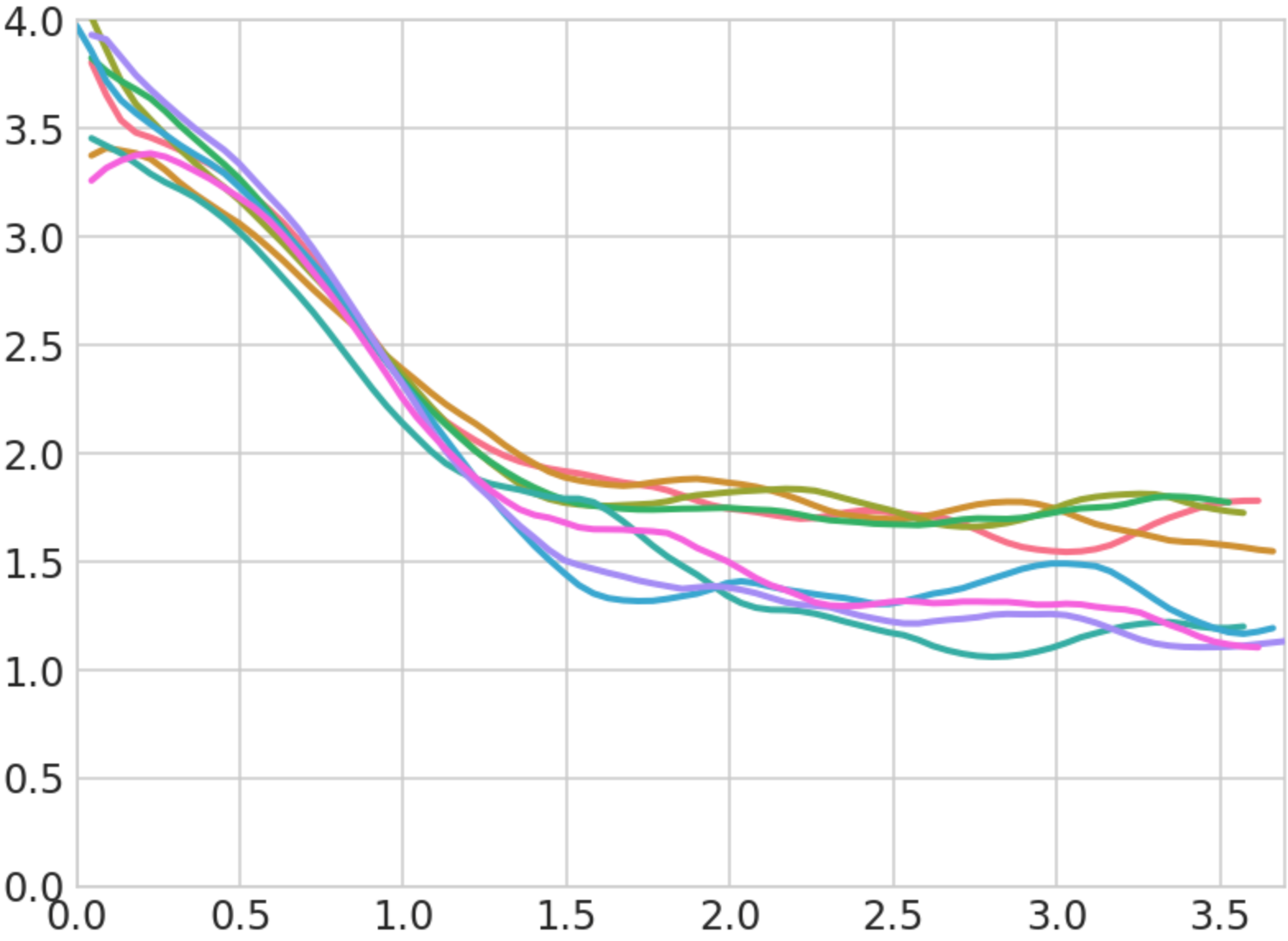}  &
&
\includegraphics[width=0.26\linewidth]{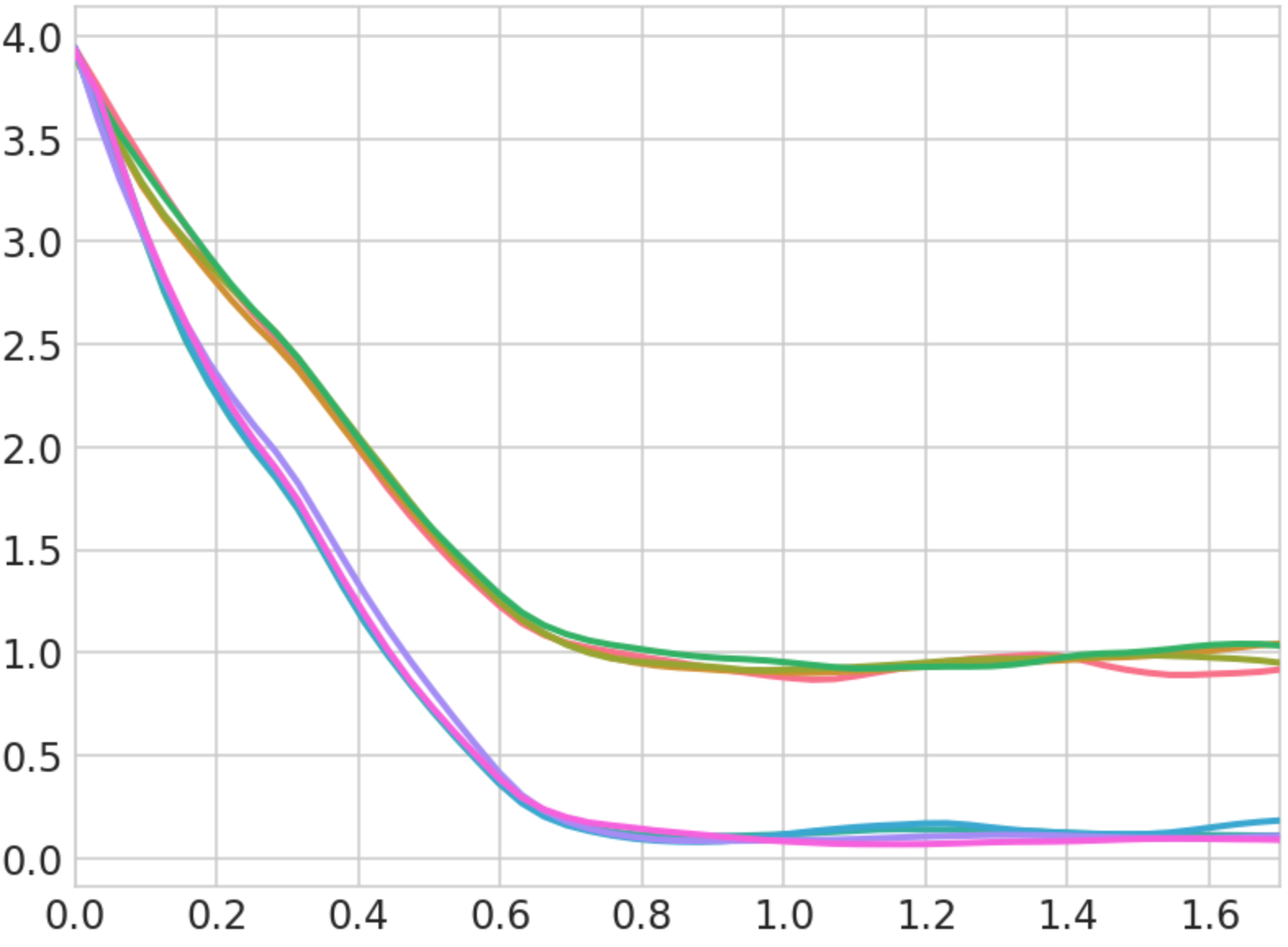} &
\includegraphics[width=0.26\linewidth]{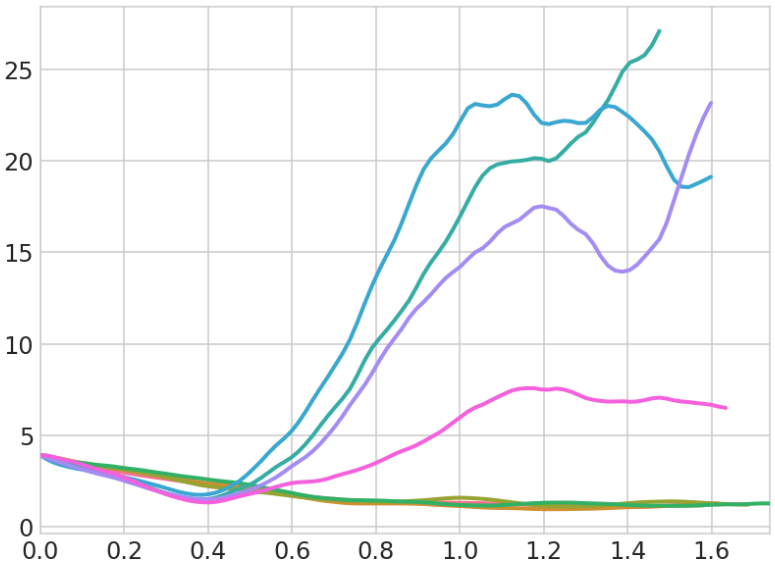} 
\\
\rotatebox{90}{\simplevgg} & 
\includegraphics[width=0.26\linewidth]{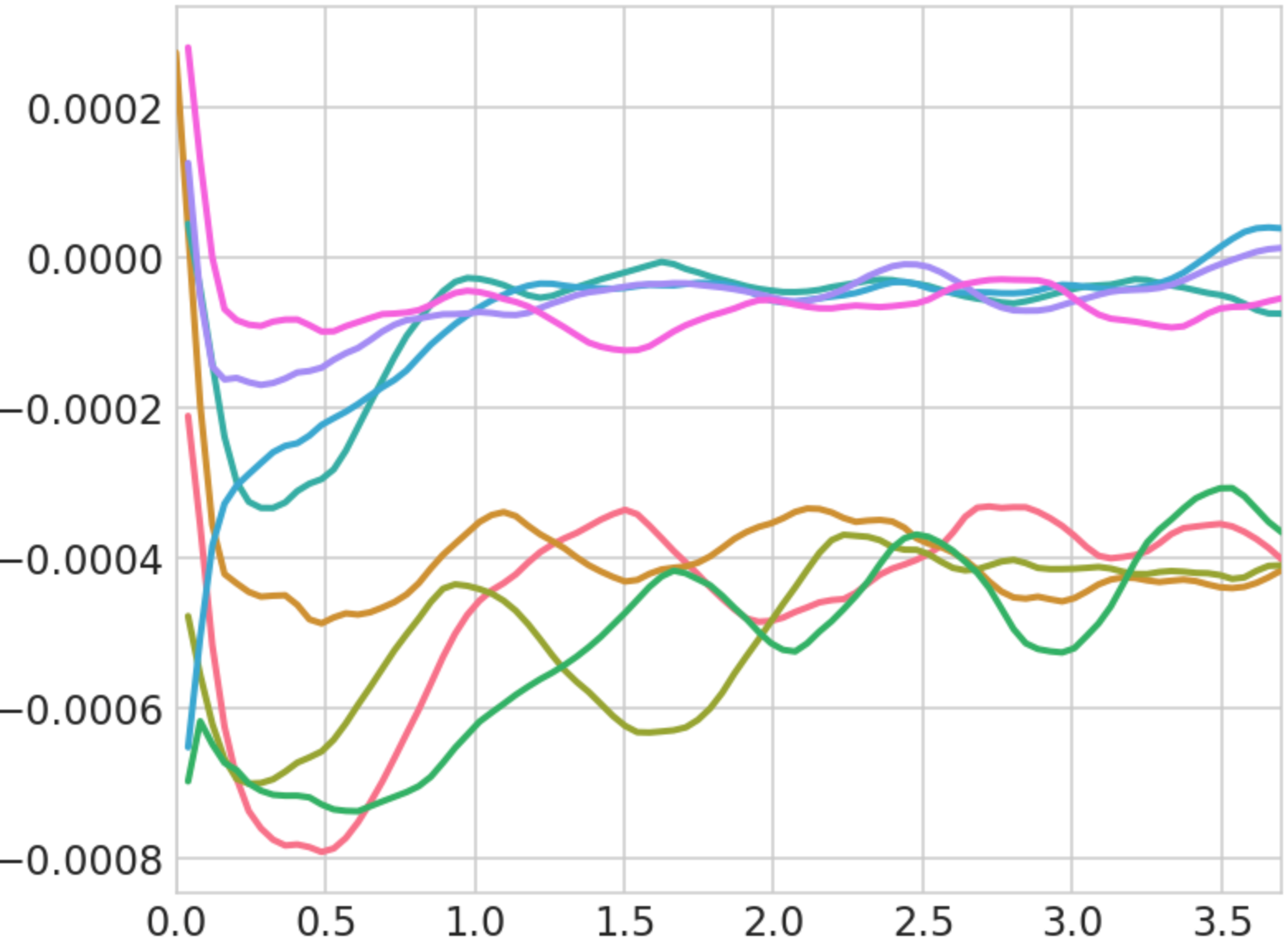} &
\includegraphics[width=0.26\linewidth]{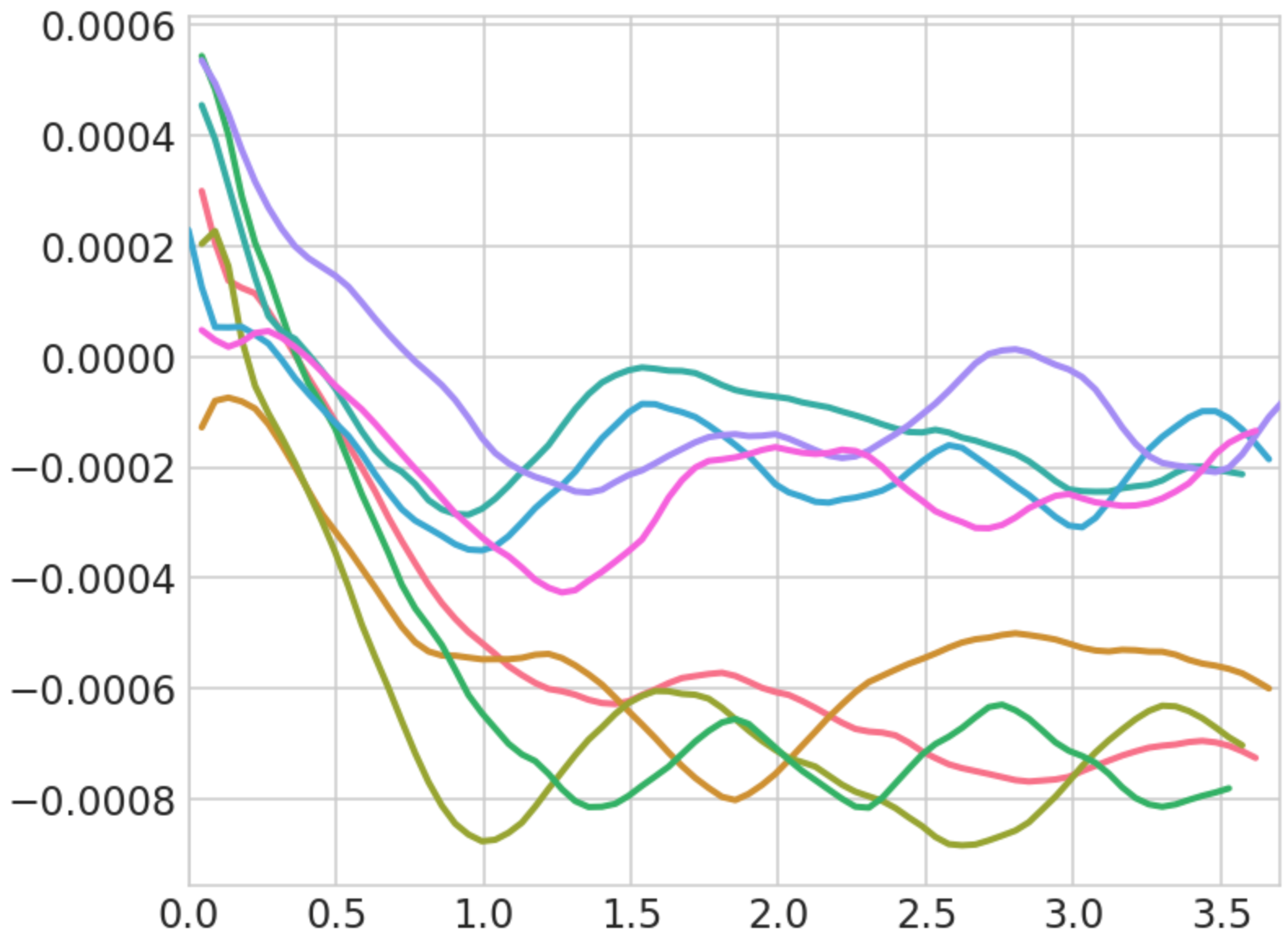}  &
\hspace{3mm}\rotatebox{90}{\simplecnn} &
\includegraphics[width=0.26\linewidth]{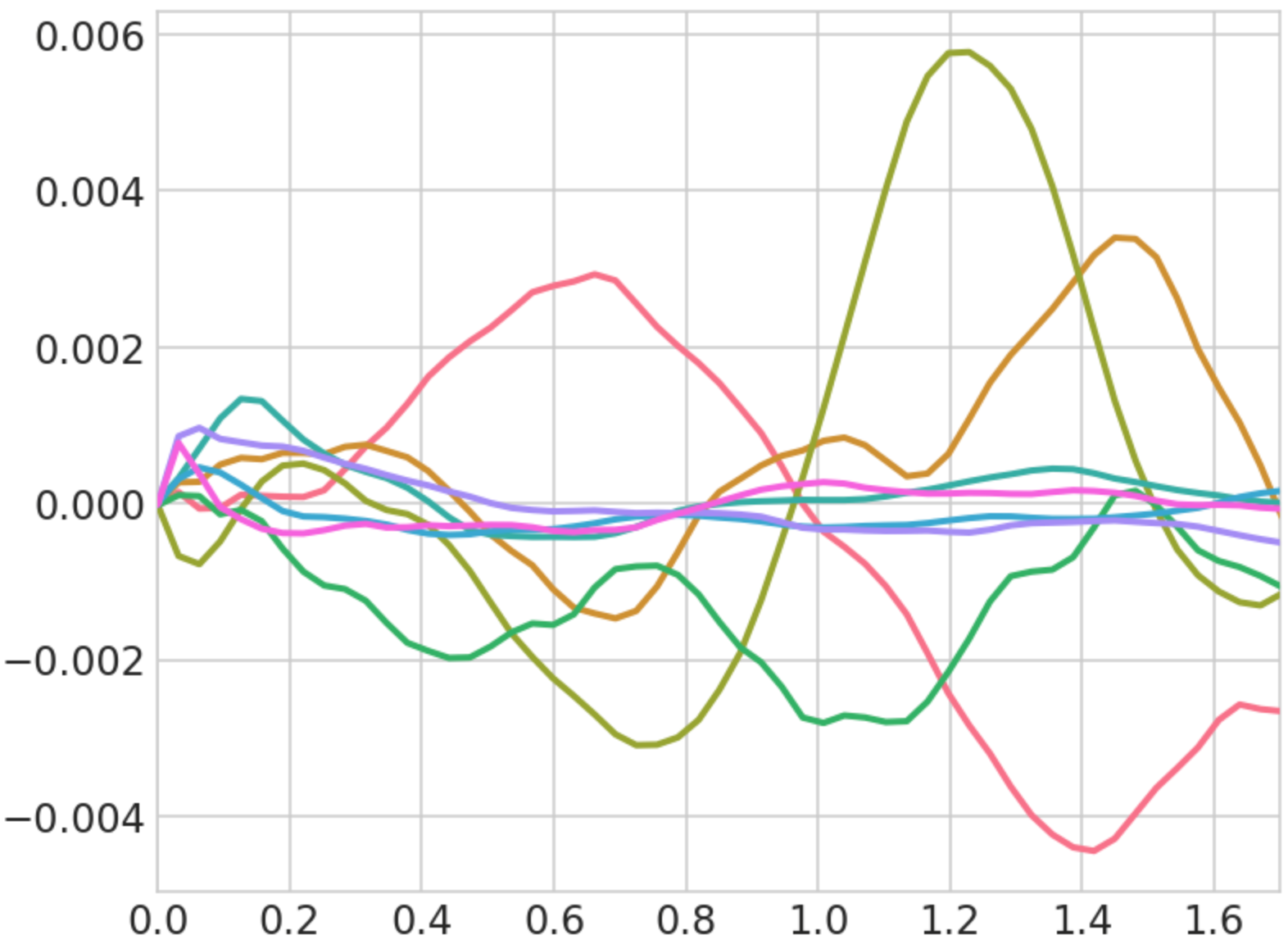} &
\includegraphics[width=0.26\linewidth]{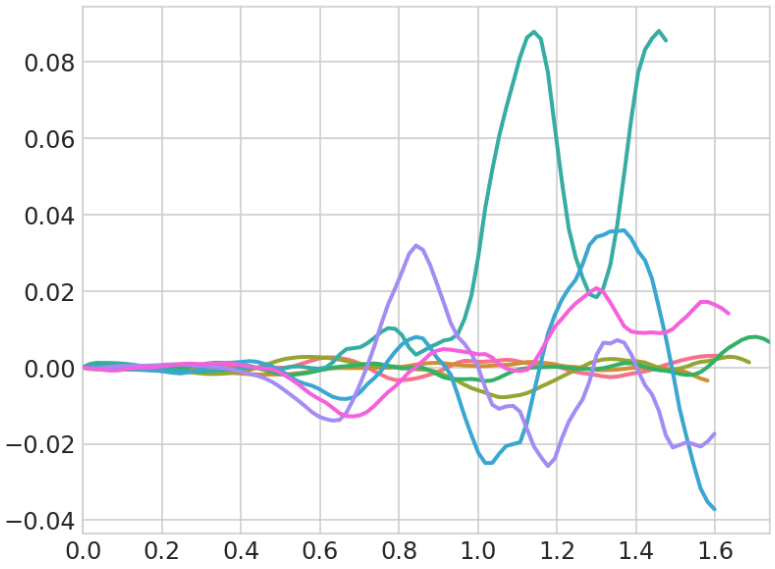} 
\\
&\includegraphics[width=0.26\linewidth]{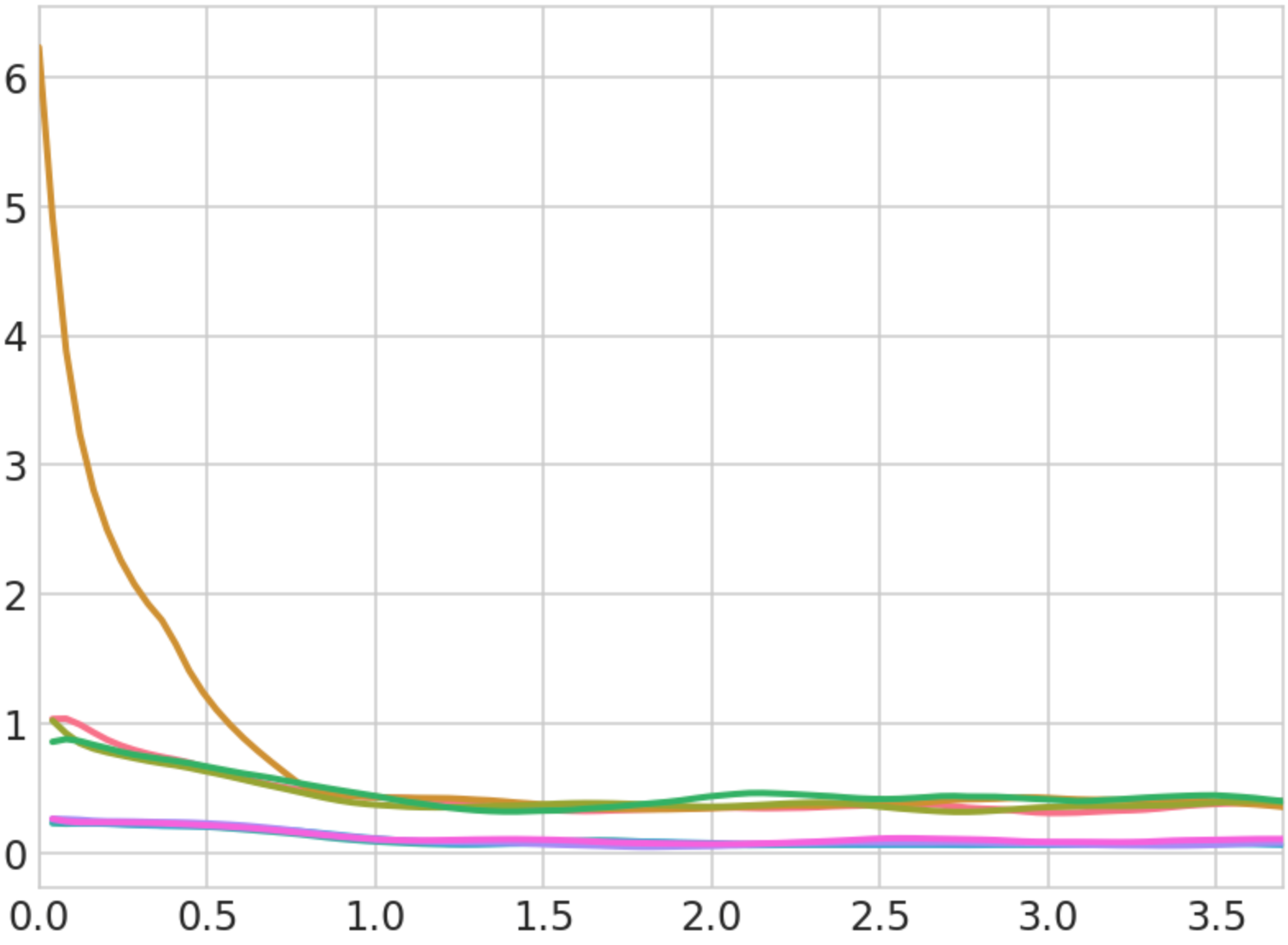} &
\includegraphics[width=0.26\linewidth]{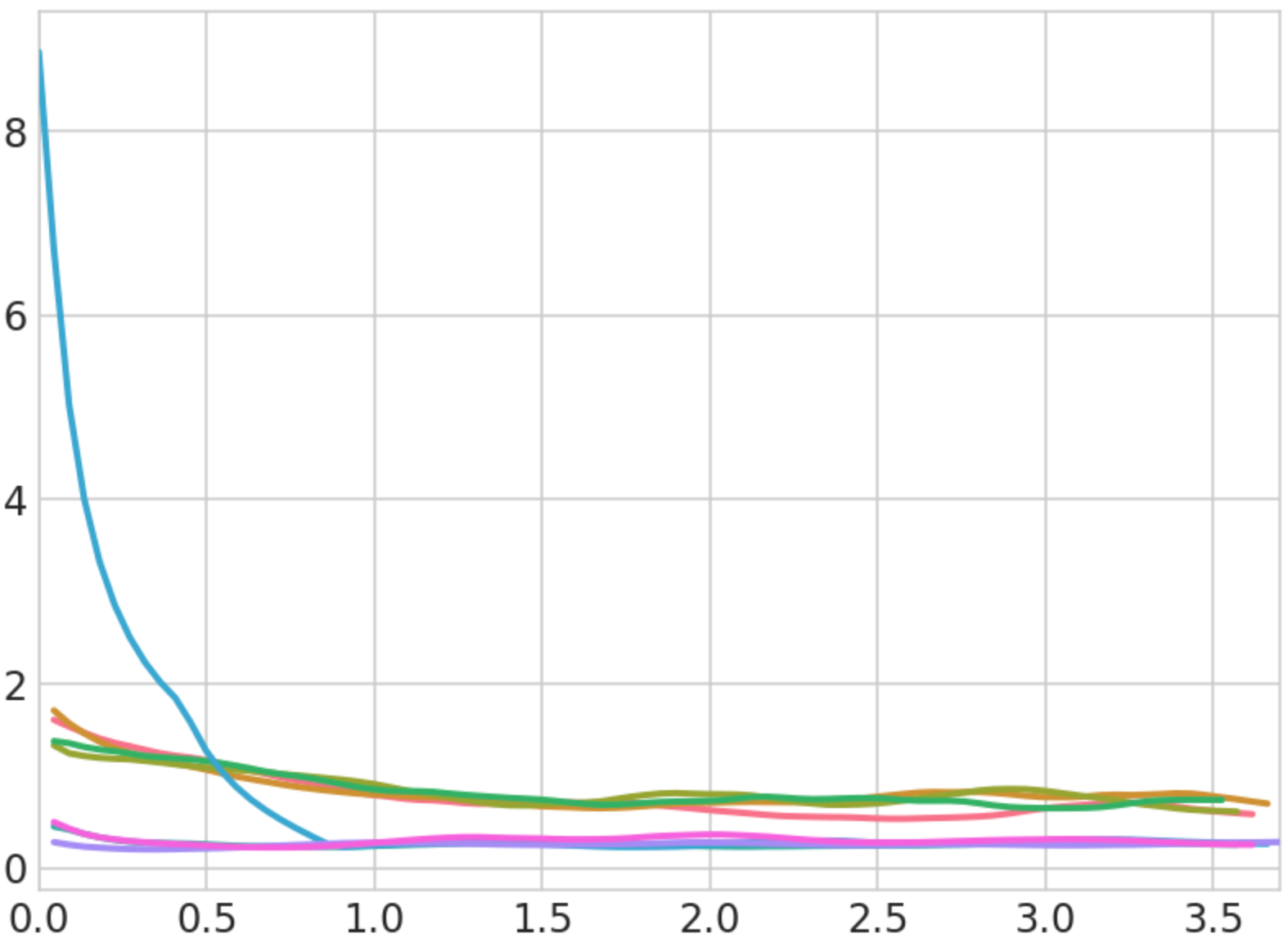}  &
&
\includegraphics[width=0.26\linewidth]{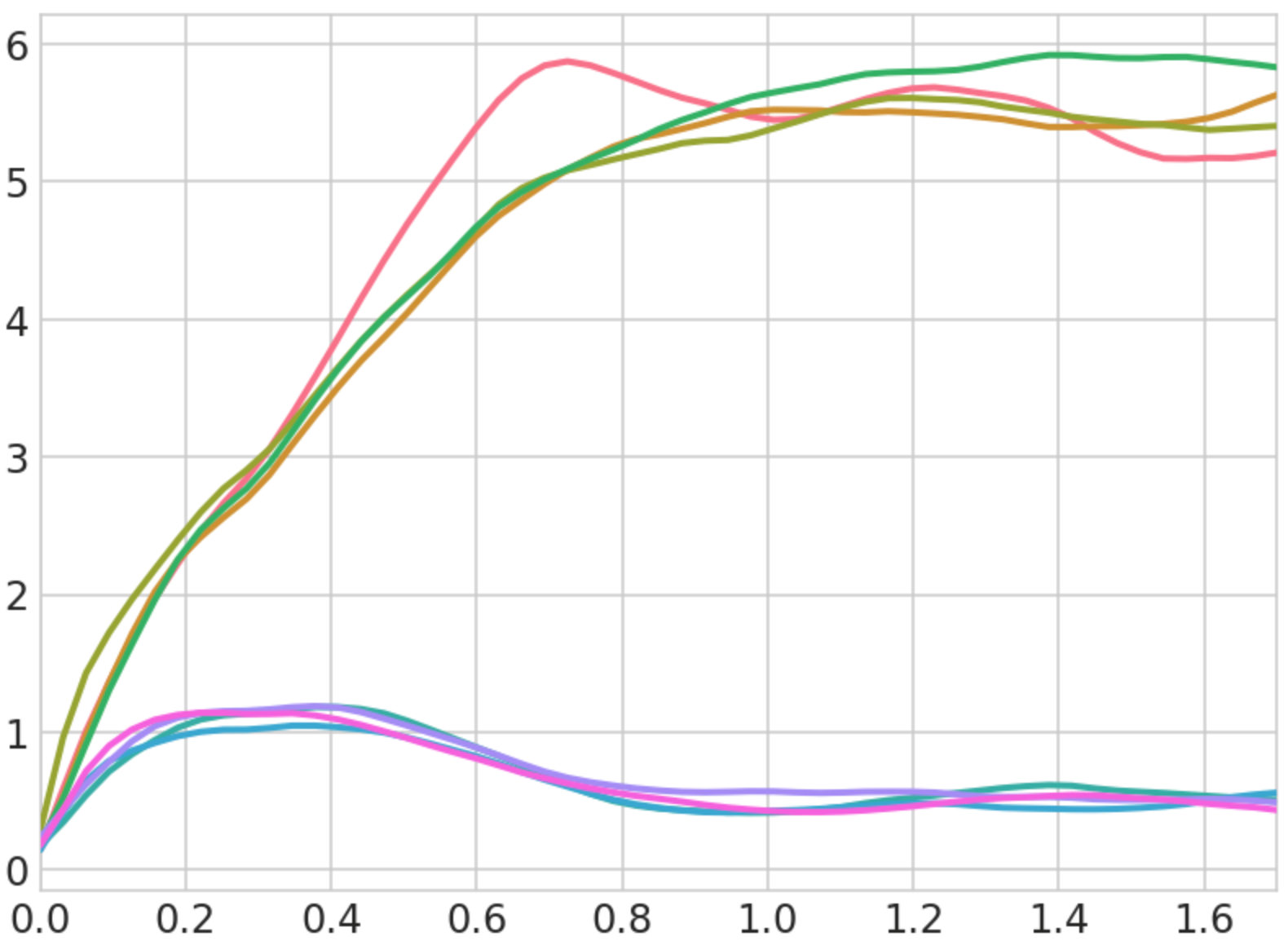} &
\includegraphics[width=0.26\linewidth]{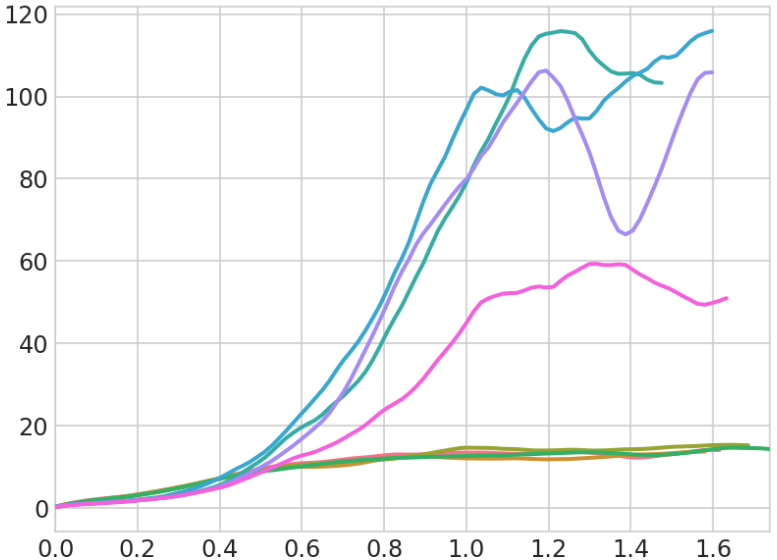} 
\\
 & \multicolumn{1}{c}{\sequential} & 
\multicolumn{1}{c}{\sideways} & 
&
\multicolumn{1}{c}{\sequential} &
\multicolumn{1}{c}{\sideways} 
\\
\bottomrule
\end{tabular}
}
\end{center}
\hspace{-1mm}
\caption{
Training dynamics of \simplecnn and \simplevgg with different models of computations. Experiments are conducted on the HMDB51 dataset. Different colors denote different hyper-parameters (red, green, olive, orange refer to the initial learning rate $10^{-5}$ and teal, pink, violet, blue to $10^{-4}$, all with various weight decay). On the x-axis, we report number of iteration steps, in $10^5$ scale. On the y-axis, we report, from the top to bottom: loss, mean of the gradients, and average gradient magnitude (l2-norm). 
Note that the figures have different y-limits to allow a detailed visualization of the training dynamics as training progresses.
}
\label{fig:training_dynamics_hmdb}
\end{figure*}
\begin{figure}[t]
\begin{center}
\scalebox{0.9}{
\begin{tabular}{c@{\ }c}
 \toprule
\includegraphics[width=0.7\linewidth]{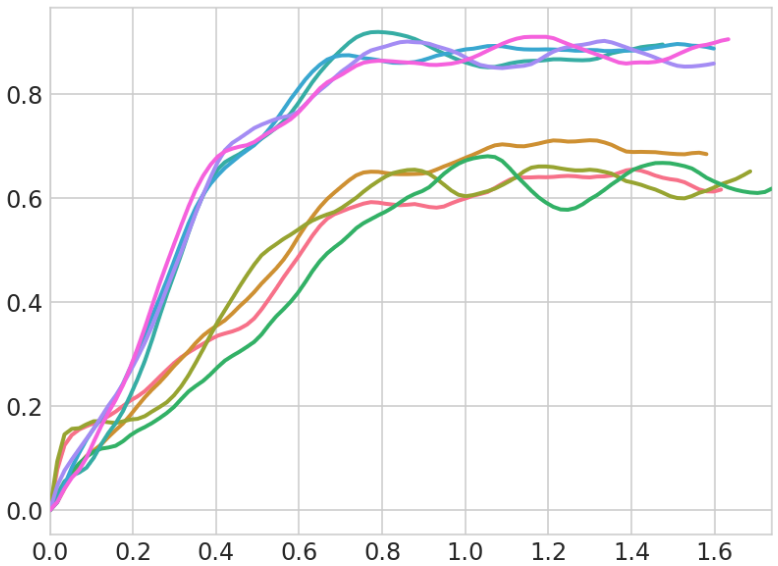} 
\\
\bottomrule
\end{tabular}
}
\end{center}
\caption{
Training of \simplecnn with the \sideways algorithm on HMDB51.
Different colors denote different hyper-parameters (the same as~\autoref{fig:training_dynamics_hmdb}). On the x-axis, 
we report number of iteration steps, in $10^5$ scale. 
On the y-axis, we report, accuracy numbers.
}
\label{fig:simple_cnn_accuracy}
\end{figure}
\begin{figure*}[t]
\begin{center}
\scalebox{0.75}{
\begin{tabular}{c@{\ }c@{\ }c@{\ }c}
 \toprule
\rotatebox{90}{\quad\quad\quad\quad favourable} & 
\includegraphics[width=0.36\linewidth]{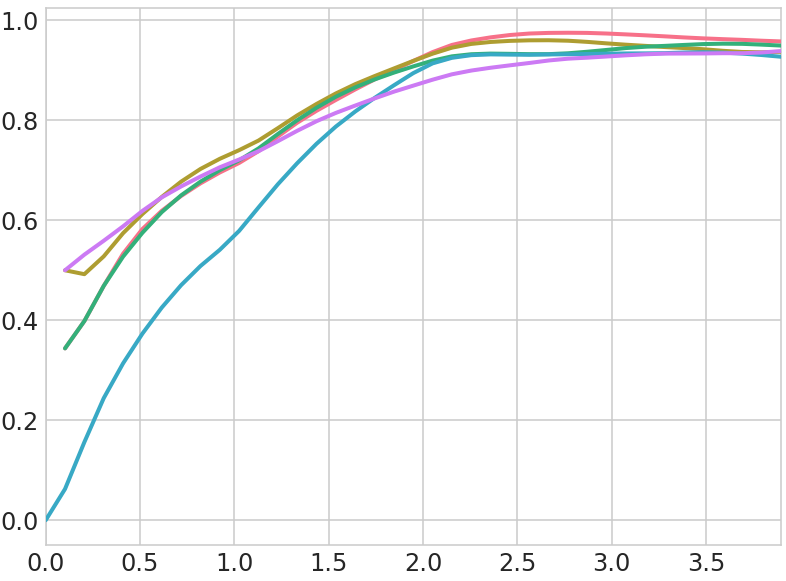} &
\hspace{5mm}
\includegraphics[width=0.36\linewidth]{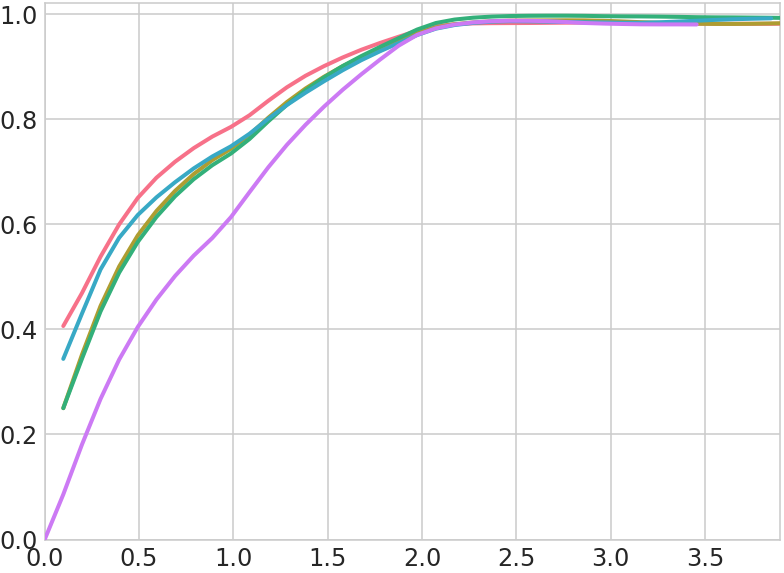}
\vspace{1mm}
\\
\rotatebox{90}{\quad\quad\quad\quad unfavourable} &
\includegraphics[width=0.36\linewidth]{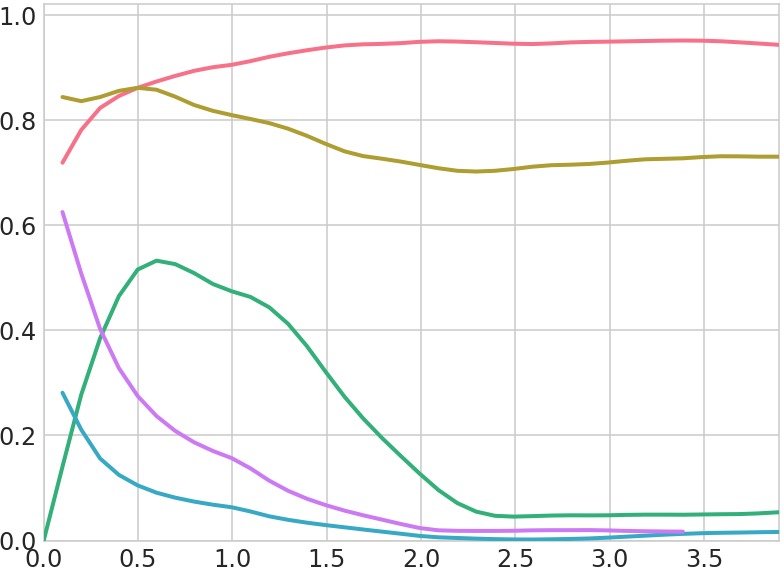} &
\hspace{5mm}
\includegraphics[width=0.36\linewidth]{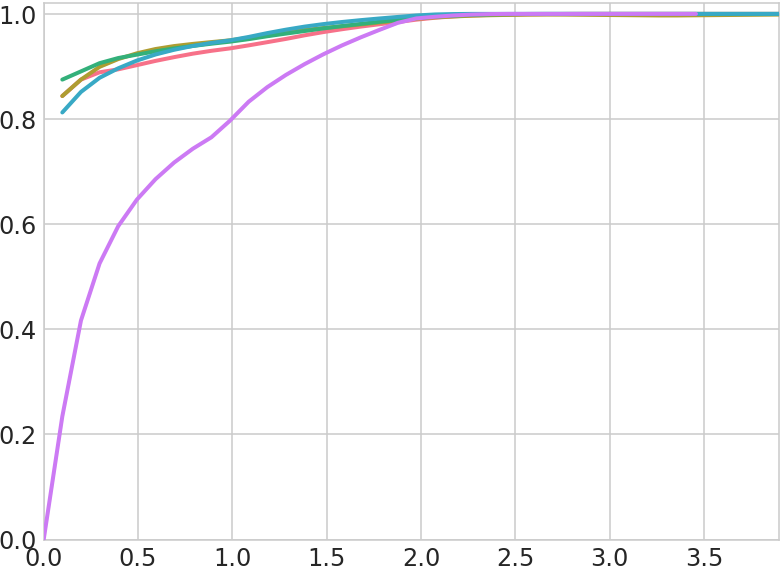}
\\
&
\multicolumn{1}{c}{\sideways} &
\multicolumn{1}{c}{\sequential} 
\\
\bottomrule
\end{tabular}
}
\end{center}
\caption{
We experiment with different temporal striding settings ($\{2,3,4,5,6\}$ encoded as red, olive, green, blue, violet, respectively) for the input videos, on UCF101. 
Top row shows favourable hyper-parameters (initial learning rate equals to $10^{-5}$). Bottom row shows unfavourable hyper-parameters (initial learning rate equals to $10^{-4}$).
First column shows \sideways. Second column shows \sequential. On the x-axis, we report number of iteration steps. On the y-axis, we report accuracy numbers.
In the setting with unfavourable hyper-parameters, training of networks collapses with higher striding numbers.  
}
\label{fig:smoothness}
\end{figure*}
\vspace{2mm}
\noindent
\textbf{Quantitative results.}
\autoref{tab:sequential_vs_sideways} directly compares both algorithms, backpropagation with the pipelined \sideways training. For the sake of comparison, we also report referenced models that are trained using `regular' training. As we can see, we have reproduced similar results with the \sequential model, and in several cases, we have achieved higher accuracy (e.g., \simplevgg + Dropout (0.9)). 
Even though higher accuracy numbers have been previously reported on both datasets, these are achieved using larger models pre-trained on larger datasets. Our focus is, however, different.

Results presented in \autoref{tab:sequential_vs_sideways} suggest that the \sideways training achieves competitive accuracy to \sequential, and the introduced noise due to i) the sampling error, the same as SGD updates, and ii) the pseudo-gradients computations, does not seem to harm the overall performance. Quite the opposite, under certain conditions, we observe \sideways  generalizes better than \sequential. Since such behavior  occurs  during training larger models on relatively small video datasets, \eg, training \simplevgg on UCF-101, which often results in overfitting~\cite{simonyan2014two}, we hypothesize \sideways acts as an implicit regularizer for video processing~\cite{neyshabur2017implicit,zhang2016understanding}. 

\vspace{2mm}
\noindent
\textbf{Training dynamics.}
Since we compare \sideways algorithm to \sequential, it is instructive to investigate their training behavior. Intuitively, similar training behavior should result in a similar final performance. Therefore, we have conducted experiments where we measure various statistics throughout training, and we report them in \autoref{fig:training_dynamics_hmdb}. There are a few interesting observations. First, the training dynamics of the \simplevgg architecture with \sideways training closely follows `regular' training (first two columns). However, for the \simplecnn architecture, training dynamics between both algorithms differ under some choice of the hyper-parameters. For instance, we can notice in \autoref{fig:training_dynamics_hmdb} (last two columns) the loss function become quite unstable. This happens consistently with a larger learning rate, \eg, above $10^{-4}$. 
Even though this seemingly should also transfer into unstable training accuracy, we have found training does not collapse. Quite the opposite, we report a relatively high training accuracy (above $85\%$) as shown in \autoref{fig:simple_cnn_accuracy}. After a more careful inspection, we observe that \simplecnn trained with \sideways and larger learning rates tends to give confident predictions that result in high loss whenever they miss the class. Results on UCF-101 are similar, but slightly less pronounced.

\vspace{2mm}
\noindent \textbf{Sensitivity to frame rate.} The smoothness of the input space is the key underlying assumption behind the \sideways algorithm. When the input space is the space of video clips, this assumption translates, \eg, into a high frame-rate. To further stretch this assumption, we have artificially decreased the frame-rate by skipping data frames in the input video clip. This can easily be implemented with the striding operation, \ie, we skip $k$ frames with striding $k+1$. To keep the length of video clips unchanged between the experiments, we sample  $k+1$ times longer input sequences, with padding, before we apply striding. We have experimented with striding in $\{2,3,4,5,6\}$. In our experiments, we have found \sideways to be surprisingly robust to the changes in striding. Only some choice of the hyper-parameters, \eg, relatively high learning rate, have resulted in the performance collapse, where the network has transitioned from high into low training accuracies. Nonetheless, \sequential  and \sideways never collapses with the same set of the carefully chosen hyper-parameters. Distortions introduced by padding could be another explanation for the collapse of models trained with `unfavorable' hyper-parameters and higher striding numbers. We report these results in \autoref{fig:smoothness}.

\vspace{2mm}
\noindent \textbf{Training speed-up using multiple GPUs.} We evaluate the speedup obtained when training the VGG-8 and \simplecnn models using a single V100 GPU per module -- 8 for VGG and 6 for \simplecnn. To isolate training speed from the data loading aspect, in this study, we artificially construct videos consisting of random numbers. We train each model for 100 steps, repeat this 3 times and return the highest average number of training steps per second. The results are shown in~\autoref{tab:speedup_sideways}, which validate that there is a large speed-up for \sideways when parallel resources are assigned along the network depth. In particular, VGG has a more balanced decomposition in terms of FLOPs per module. The \sequential model benefits little from the multiple GPUs since they are locked most of the time waiting for the other GPUs to complete their processing. Note also, that placing different \sideways modules in different GPUs will also significantly reduce memory requirements for training large neural networks.

\begin{table}[t]
\small
    \centering
    \resizebox{0.9\linewidth}{!}{%
    \begin{tabular}{llll}
        \toprule
        &  \sequential & \sideways & speedup  \\
        \midrule
        Simple CNN & 1.7 & 8.4 & 4.9x \\
        VGG-8   & 0.1 &  0.6 & 6.0x \\
        \bottomrule 
    \end{tabular}}
    \caption{Number of training steps per second for two architectures, using batch size of 8 clips, each having 64 frames and resolution 112x112. The results were obtained using one GPU per network module (6 for Simple CNN and 8 for VGG-8).}
    \label{tab:speedup_sideways}
\end{table}

\subsection{Results (Auto-Encoding)}
\begin{figure}[tbh]
\begin{center}
\begin{tabular}{c@{\ }c}
\includegraphics[width=0.97\linewidth]{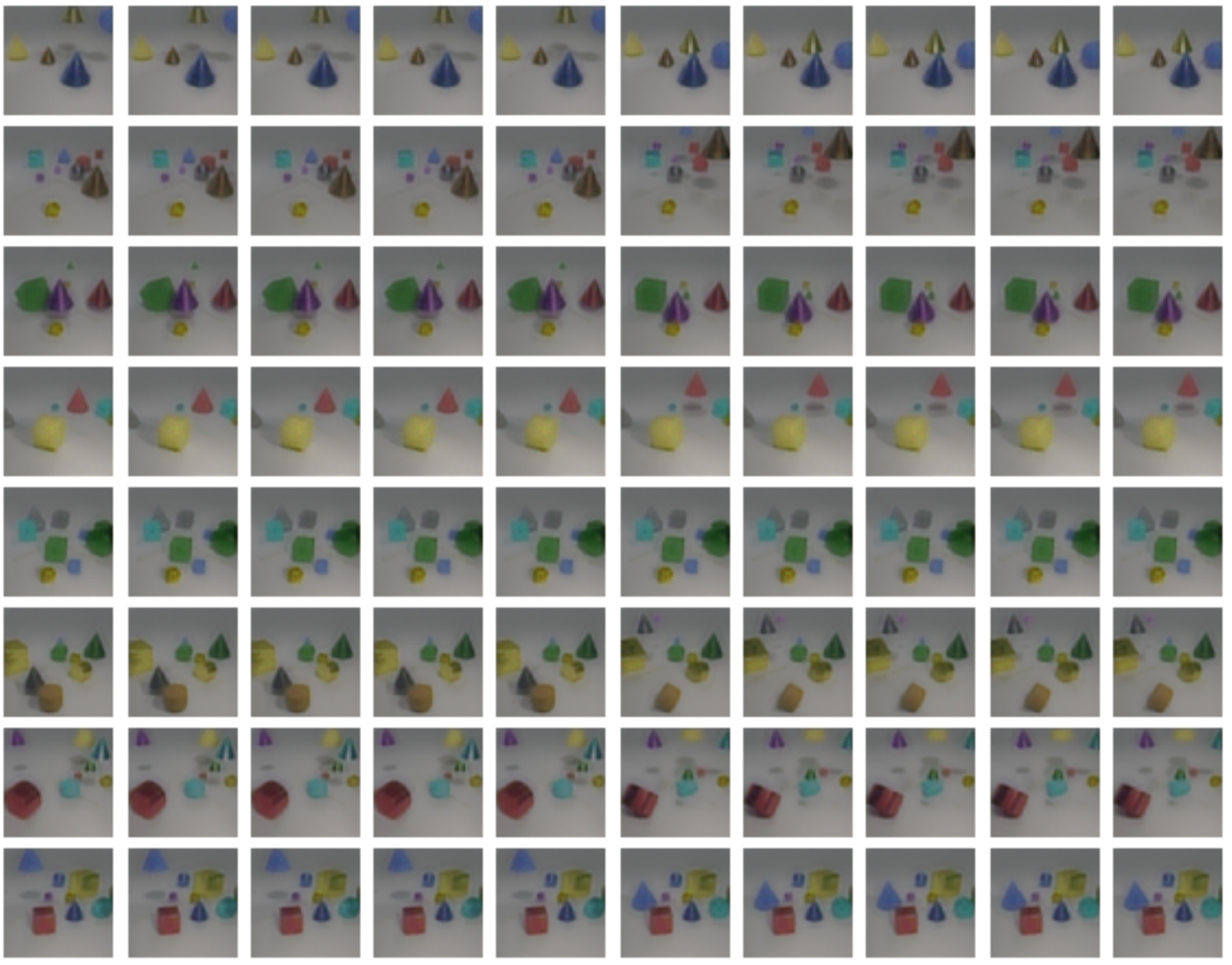}
\end{tabular}
\end{center}
\caption{
Auto-encoding results for \sequential. Each row shows an individual video sequence consisting of, in our case, $64$ frames. For the sake of visualization, we sub-sample $10$ consecutive middle frames from the outcome.
}
\label{fig:sequential_predictions}
\end{figure}

\begin{figure}[H]
\begin{center}
\begin{tabular}{c@{\ }c}
\includegraphics[width=0.98\linewidth]{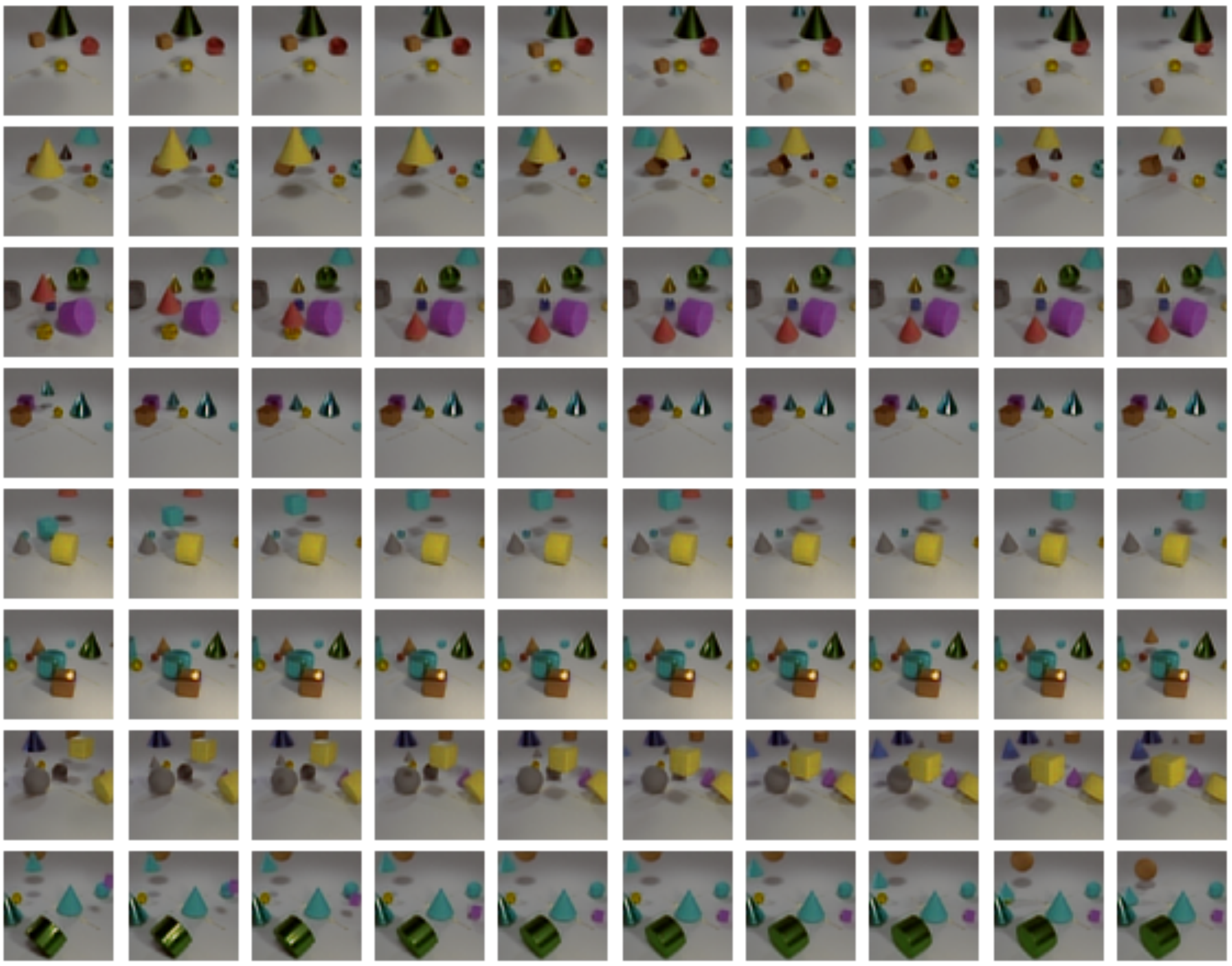}
\end{tabular}
\end{center}
\caption{
Auto-encoding results for \sideways. Each row shows an individual video sequence consisting of, in our case, $64$ frames. To simplify visualization, we sub-sample $10$ consecutive middle frames from the outcome.
}
\label{fig:sideways_predictions}
\end{figure}

We evaluate both algorithms using mean squared error, between pixels, between the predicted and the ground truth sequences, under the same conditions, in particular, under the same frame rate. 
\noindent
\vspace{2mm}
\newline
\textbf{Qualitative results.}
We show qualitative results of the \sequential outputs in \autoref{fig:sequential_predictions} and \sideways outputs in \autoref{fig:sideways_predictions}. Shapes, colors, as well as many fine-grained attributes such as metallic materials are properly decoded. We have found that the model trained with \sequential is successful at decoding coarser objects attributes, but is slightly less accurate with metallic materials. As mentioned above, because of the blocking mechanism of the \sequential algorithm and the high frame rate of the input, the method needs to discard input frames to not accumulate latency. In these cases, the last produced output is copied to compensate for the low output rate, resulting in identical output frames at different time steps over some time interval -- and different ones at the beginning of the next \updatecycle (this is best visible in the 1st row of \autoref{fig:sequential_predictions}).
\noindent
\vspace{2mm}
\newline
\textbf{Quantitative results.}
\autoref{tab:mse} shows mean squared error (the lower, the better) between predicted frames and ground truth (input frames). We compare the same architecture trained with \sideways and the regular \sequential. Because of the synchronization, the method trained with \sequential cannot output at a fast enough pace to keep up with the input frame rate, which yields a significant error. 
\begin{table}[tbh]
\small
    \centering
    \resizebox{0.75\linewidth}{!}{%
    \begin{tabular}{lll}
        \toprule
               &  \sequential & \sideways  \\
        \midrule
        Auto-encoding & $0.014$ & $0.002$  \\
        \bottomrule 
    \end{tabular}}
    \caption{Mean squared error between predictions and ground truth data; the lower, the better.}
    \label{tab:mse}
\end{table}

These results show that the proposed training scheme can be  successfully applied also to tasks where both input and output are continuously evolving, reducing considerably the latency of the system during training. Together with the results on the classification task, we conclude that this training scheme is general enough to be applied for a wide range of video tasks. 

\section{Conclusion}

We have proposed \sideways\;-- a backpropagation variant to train networks, where activations from different \timesteps are used in the weight updates.
We motivate our training algorithm by the smoothness of video signals, and especially we assume that important features vary slowly, at least in the latent space~\cite{hinton1990connectionist,wiskott2002slow}.

We have found that \sideways is not only a valid learning mechanism but can also potentially provide an implicit regularization during the training of neural networks. Decoupling provided by the \sideways algorithm makes it especially attractive for training large models in parallel.

We hope that our work will spark further interest in developing decoupled training of more advanced temporal models or in a better understanding of the role of slow features, temporal redundancies, and stochasticity in the learning process of such models. 
Although biological plausibility is not our primary motivation, we believe our architecture has some desired properties. For instance, top-down and global communication implemented in \sideways does not necessarily require neither depth-synchronization nor instantaneous propagation; it also does not require local caching of the activations during the weights updates for the backward pass. Finally, its unrolling in time could be viewed as more biologically correct~\cite{betti2019backprop,kubilius2018cornet}.

\bibliographystyle{abbrvnat}
\setlength{\bibsep}{4pt} 
\setlength{\bibhang}{0pt}
\bibliography{refs}

\section*{Acknowledgements}
We thank Jean-Baptiste Alayrac, Carl Doersch, Tamara Norman, Simon Osindero, and Andrew Zisserman for their critical feedback, and their help during the duration of this project.

\end{document}